\documentclass[conference]{IEEEtran}
\IEEEoverridecommandlockouts
\usepackage{cite}
\usepackage[colorlinks,
            linkcolor=black,
            anchorcolor=black,
            citecolor=black,
            urlcolor=black]{hyperref}
\usepackage{amsmath,amssymb,amsfonts}
\usepackage{graphicx}
\usepackage{textcomp}
\usepackage{xcolor}
\usepackage{colortbl}
\usepackage{enumitem}
\usepackage{pifont}
\usepackage{booktabs}
\usepackage{makecell}
\usepackage{multirow}
\usepackage{amsmath}
\usepackage{subfigure}
\usepackage{balance}
\usepackage{algorithmic}
\usepackage{footmisc}
\usepackage{algorithm}

\usepackage{bbm}
\def\BibTeX{{\rm B\kern-.05em{\sc i\kern-.025em b}\kern-.08em
    T\kern-.1667em\lower.7ex\hbox{E}\kern-.125emX}}
\begin{document}

\title{From Chaos to Clarity: Time Series Anomaly Detection in Astronomical Observations \\
% Detecting Anomalies in Astronomical Light Curves Eliminating Concurrent Noise\\
\thanks{\noindent \footnotesize{\# Corresponding author: Xiaofeng Meng.} }
}

% \author{
%     \IEEEauthorblockN{Xinli Hao$^{1}$, XX$^{a,b}$, XX$^b$}
%     \IEEEauthorblockA{$^a$ Renmin University of China, Beijing, China}
%     \IEEEauthorblockA{$^b$ Department of Computer Science and Technology, Tsinghua University, Beijing, China}
%     \IEEEauthorblockA{\{xinli\_hao\}@ruc.edu.com, \{XX, XX\}@XXX.edu.cn}
% }

\author{\IEEEauthorblockN{Xinli Hao\IEEEauthorrefmark{1}, Yile Chen\IEEEauthorrefmark{2}, Chen Yang\IEEEauthorrefmark{3}, Zhihui Du\IEEEauthorrefmark{4}, Chaohong Ma\IEEEauthorrefmark{1}, Chao Wu\IEEEauthorrefmark{5},Xiaofeng Meng\#\IEEEauthorrefmark{1}}

\IEEEauthorblockN{\IEEEauthorrefmark{1}Renmin University of China, Bejing, China\\
\IEEEauthorrefmark{2}Nanyang Technological University, Singapore\\
\IEEEauthorrefmark{3}China National Clearing Center, Beijing, China\\
\IEEEauthorrefmark{4}New Jersey Institute of Technology, Newark, USA\\
\IEEEauthorrefmark{5}National Astronomical Observatories, CAS, Beijing, China}
% \IEEEauthorblockA{\email{chaohma, xfmeng, aishan@ruc.edu.cn}, \email{xhyu@yorku.ca}, \email{yifanli@eecs.yorku.ca} }
\{xinli\_hao, chaohma, xfmeng\}@ruc.edu.cn, yile001@e.ntu.edu.sg, \\
chenyang@cncc.cn, zhihui.du@njit.edu, cwu@bao.ac.cn 

% <-this % stops an unwanted space
% \IEEEauthorblockA{\href{mailto:chaohma@ruc.edu.cn}{chaohma}@ruc.edu.cn, \{\href{mailto:lisi@XXX.edu.cn}{lisi}, \href{mailto:wangwu@XXX.edu.cn}{wangwu},\}@XXX.edu.cn, \href{mailto:g.li@XXX.com}{g.li}@XXX.com} 
}

\maketitle

\begin{abstract}
With the development of astronomical facilities, large-scale time series data observed by these facilities is being collected. Analyzing anomalies in these astronomical observations is crucial for uncovering potential celestial events and physical phenomena, thus advancing the scientific research process. However, existing time series anomaly detection methods fall short in tackling the unique characteristics of astronomical observations where each star is inherently independent but interfered by random concurrent noise, resulting in a high rate of false alarms. To overcome the challenges, we propose AERO, a novel two-stage framework tailored for unsupervised anomaly detection in astronomical observations. In the first stage, we employ a Transformer-based encoder-decoder architecture to learn the normal temporal patterns on each variate (i.e., star) in alignment with the characteristic of variate independence. In the second stage, we enhance the graph neural network with a window-wise graph structure learning to tackle the occurrence of concurrent noise characterized by spatial and temporal randomness. In this way, AERO is not only capable of distinguishing normal temporal patterns from potential anomalies but also effectively differentiating concurrent noise, thus decreasing the number of false alarms. We conducted extensive experiments on three synthetic datasets and three real-world datasets. The results demonstrate that AERO outperforms the compared baselines. Notably, compared to the state-of-the-art model, AERO improves the F1-score by up to 8.76\% and 2.63\% on synthetic and real-world datasets respectively. 
% In the first stage, we employ a Transformer-based encoder-decoder architecture on each variate in alignment with the characteristic of object independence.
% However, existing time series anomaly detection methods fall short in tackling the unique characteristics of astronomical observations in which each star is independent but interfered with by correlated yet random concurrent noise, resulting in a high rate of false alarms.
% We publish the code and datasets on GitHub for better reproducibility.
\end{abstract}


\section{Introduction}
\label{section:1}
\noindent 
% With the development of large-scale data collection devices, it has become possible to monitor a large number of objects simultaneously. Examples include large telescopes for celestial observations and remote sensing satellites for earth observations. A single device can continuously monitor multiple objects that are within a specific area, thereby generating a special data stream, called catalog stream(which will be introduced in the next paragraph), as shown in Fig \ref{fig:catalog}. Offline mining of this kind of data can reveal important scientific laws or analyze urban conditions, and real-time anomaly detection enables timely discovery of significant scientific phenomena, major disasters, and so on. In this study, we focus on anomaly detection analysis of the data: improving online monitoring performance through offline training. 

In recent years, scientific discovery in astronomy has experienced significant advancements owing to the development of modern facilities, such as optical or radio telescopes with higher temporal and spatial resolution. These cutting-edge developments have greatly facilitated the collection of large-scale astronomical data, including optical images and radio signals, thus providing researchers with valuable resources to explore and explain the natural world. Meanwhile, the availability of such extensive datasets highlights the necessity for automated data analysis and interpretation to enhance related disciplines. In particular, anomalies (i.e., deviations from the normal patterns) identified in astronomical observations are critical indicators for potential celestial events or physical phenomena\cite{catalog}. Given the vast volume of data records, it is important to design an effective anomaly detection method to assist in the scientific research process.
% catalog streams

\begin{figure}[t]
\includegraphics[width=0.98\columnwidth]{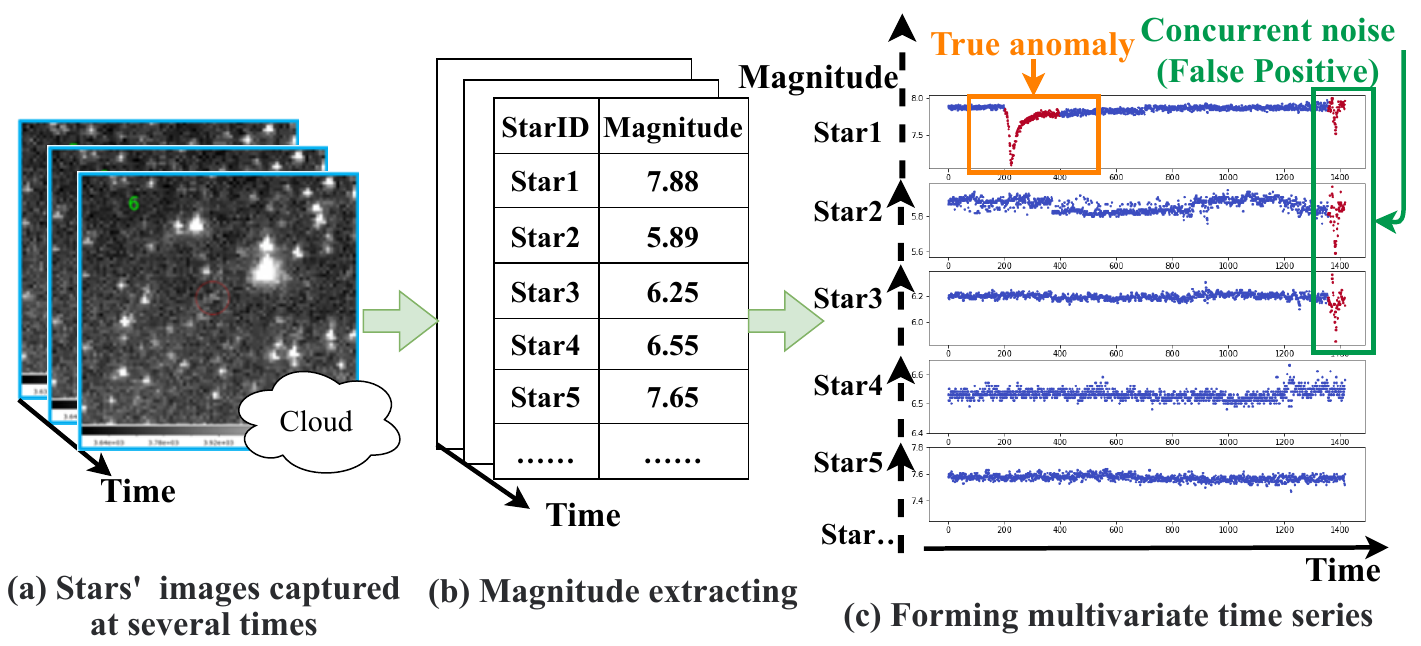} 
\vspace{-3mm}
\caption{An illustration for multivariate time series obtained from multiple stars and examples for anomaly detection. (a) Images of stars captured by telescopes. (b) The magnitudes (i.e. brightness) of stars are extracted. (c) Multiple magnitudes series constitute a multivariate time series containing true anomalies and concurrent noise. }
% multiple stars are disturbed by the same noise at the same time, namely concurrent noise. This kind of noise decreases the anomaly detection performance for many anomaly detection methods.

\vspace{-3mm}
\label{fig:catalog}
\end{figure}

% In the catalog stream, each element is no longer a single value, but a catalog containing information from multiple objects\cite{catalog}. For example, as shown in Fig \ref{fig:catalog}, an image reflecting the brightness information of numerous stars can be transformed into a catalog containing brightness information of multiple celestial bodies. In this way, an image stream can be transformed into a catalog stream. Furthermore, extracting brightness from the view of a single object can form a multivariate time series. \textbf{In this way, the catalog stream is transformed into a special kind of multivariate time series.}

The collected astronomical observations of multiple stars can be represented as a kind of multivariate time series. 
For example, as depicted in Fig.~\ref{fig:catalog}, the magnitude (i.e. brightness) of a star can be extracted from the image observed by telescopes. The magnitudes of a given star from a sequence of observations form a univariate time series. Subsequently, by aggregating multiple univariate time series, a multivariate time series can be constructed, which serves as the foundational data format for further analysis and scientific event discovery.
% These magnitudes, when concatenated by objects and aggregated over time, manifest as multivariate time series.
% These data streams, when aggregated over time, manifest as multivariate time series.
% Such aggregation serves as the foundation for monitoring abnormal celestial behaviors, thereby contributing to the subsequent scientific analysis. 
While sharing the general format of multivariate time series, the data obtained from astronomical observations possess unique properties: 
% 数据特点
% Different from general multivariate time series, the catalog stream possesses the following \textbf{three characteristics}, bringing challenges for anomaly detection of the data:
\textit{1) Variate independence.}
In astronomical multivariate time series, each variate corresponds to the time series of a star's magnitude. Given that stars are separated by vast physical distances and characterized by various physical properties, these variates are considered inherently independent and lack mutual influence. This contrasts with the multivariate time series from industrial devices\cite{FuSAGNet} or IT systems\cite{InterFusion}, where variates, typically from a single entity such as a server, exhibit close interdependencies throughout their operation;
\textit{2) Concurrent noise.} Astronomical observations obtained via telescopes are vulnerable to environmental interferences, such as shadowing effects from cloud coverage or extreme weather conditions. 
When affected by such factors, the magnitudes of a subset of stars exhibit simultaneous fluctuations over a time period, a phenomenon characterized as concurrent noise (Fig.~\ref{fig:catalog}(c)). Moreover, the presence of this concurrent noise is both spatially and temporally random, rendering its occurrence inherently unpredictable.
% \textcolor{blue}{These factors could induce noise that affects the observed magnitudes of some stars simultaneously in a period, referred to as concurrent noise (Fig.~\ref{fig:catalog}(c)). 
However, such noise does not correspond to actual celestial events, and therefore should not be interpreted as true anomalies. 
% \textbf{In a word, variables are usually independent of each other, with correlations emerging only when concurrent noise occurs}. 
Unfortunately, these two unique properties present challenges to existing time series anomaly detection methods, because they are not designed to accommodate these two characteristics. 
% and can not distinguish true anomaly and concurrent noise from anomaly candidates.

% 
% \textit{1)Object inherently independent.} The multiple objects within a catalog are inherently independent. Since every star is an independent unit having its own physical property, the brightness is independent of each other intrinsically and does not affect each other.
% \textit{2)Noise occurred dependently.} The multiple objects are disturbed by a common source of noise at the same time, which is called \textit{\textbf{concurrent noise}}. Due to device and environmental factors, multiple objects correlate together by noise. For example, a telescope malfunction can cause an increase in the brightness of all celestial bodies at the same time.
% \textit{3)Noise occurred randomly.} The correlations resulting from device and environmental factors exhibit temporal and spatial randomness. For instance, the passing of a cloud can lead to a decrease in the brightness of certain celestial bodies within a random region and at random times.

\begin{figure}[t]
\includegraphics[width=1\columnwidth]{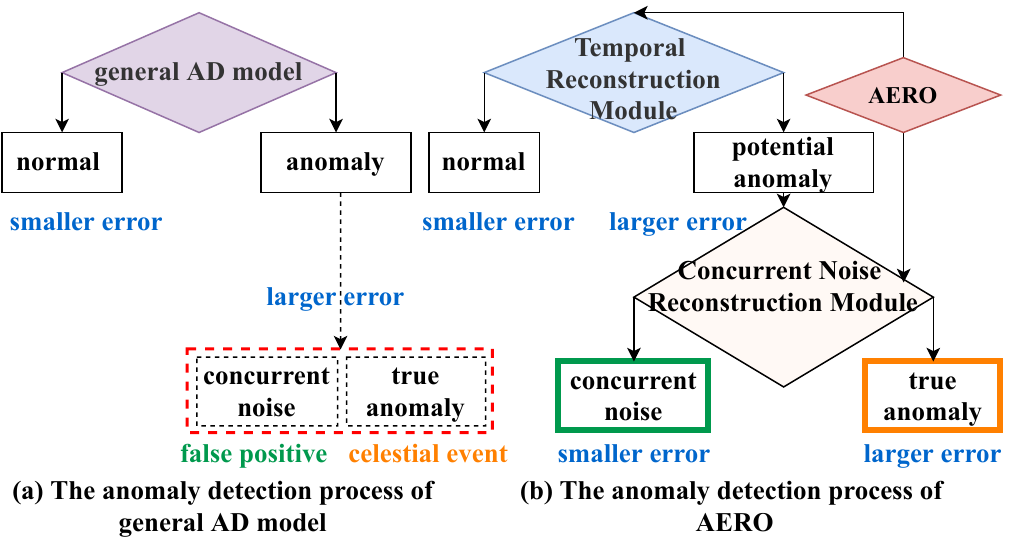} % Reduce the figure size so that it is slightly narrower than the column.
\vspace{-5mm}
\caption{Illustration of the anomaly detection process. (a) existing anomaly detection (AD) methods fail to distinguish normal data from concurrent noise in astronomical observations.   (b) AERO overcomes the limitations through a two-stage framework.}

\label{fig:flow}
\vspace{-3mm}
\end{figure}

% However, due to the lack of consideration of the above characteristics, general methods are not effective in anomaly detection of such data. As shown in Fig \ref{fig:catalog} (c) and Fig \ref{fig:flow} (a), they mistakenly identify concurrent noise as anomalies according to large reconstruction or prediction error, resulting in a high number of false positives and lower precision. Therefore, \textbf{for anomaly detection accurately on catalog streams, the model must further distinguish true anomaly and concurrent noise, thereby reducing false positives}. In other words, concurrent noise should get a small error by further modeled and leave the true anomaly a large error, as shown in Fig \ref{fig:flow} (b). There are two major challenges to this idea.
% 加参考文献
First, to tackle the property of variate independence, it is natural to employ univariate time series anomaly detection methods, which focus on modeling each variate separately\cite{surveyAnomaly}. While these methods are well-suited to the aspect of variate independence, they fail to identify the concurrent noise due to their inability to capture correlations across multiple stars. As a result, such noise might be mistakenly classified as true anomalies (celestial events), leading to an increased rate of false positives (Fig.~\ref{fig:flow}(a)). Conversely, to tackle the property of concurrent noise, it is necessary to treat the magnitudes series of multiple stars jointly as a multivariate time series. However, existing methods for multivariate time series anomaly detection are primarily designed for systems, such as industry devices and urban infrastructures\cite{GTA,FuSAGNet,MSCRED}. They operate under the assumption of persistent and predictable correlations among all variates. Such an assumption contradicts the inherent independence characteristic among stars during the absence of concurrent noise. 
% Furthermore, these methods are inadequate for modeling the unpredictable spatial and temporary randomness associated with concurrent noise. 
Therefore, these methods result in suboptimal performance when applied to astronomical observations. 

Second, recent studies have proposed to treat each variate in multivariate time series as a node, and applied graph neural networks (GNNs) to uncover latent dependencies among these variates\cite{GDN,METRO,Event2Graph}. While these methods demonstrate enhanced performance in capturing latent dependencies among variates compared to conventional multivariate time series methods, they still encounter issues in effectively modeling concurrent noise. Specifically, GNNs in these methods are developed to explicitly learn the structure of either a global static graph or a dynamic temporal graph. As mentioned previously, the intrinsic unpredictability of environmental interferences indicates that concurrent noise exhibits both spatial and temporal randomness. In other words, concurrent noise can affect an undetermined number of stars within any region while occurring at any time. This suggests the absence of stable dependencies among variates and the lack of predictable, fixed temporal evolution patterns.
Such randomness not only undermines the basic premise of constructing a static graph, but also challenges the core rationale behind the implementation of learning a dynamic graph.

To overcome these challenges, we propose a novel method named AERO for \textbf{A}nomaly d\textbf{E}tection in ast\textbf{R}onomical \textbf{O}bservations in an unsupervised manner. 
Our method employs a two-stage detection framework to address the limitations in both types of existing anomaly detection methods. In the first stage, we adopt a Transformer-based encoder-decoder model, and apply it independently to each variate to reconstruct univariate time series. This strategy encodes prior knowledge of variate independence, thus enabling the model to learn the normal temporal patterns of stars effectively. Through this process, potential anomalies with large reconstruction errors can be preliminarily identified. In the second stage, based on the results from the first stage, we consider the variate dependencies to further differentiate true anomalies and concurrent noise, as shown in Fig.~\ref{fig:flow}(b). 
% The dependency modeling approach diverges from existing GNN-based anomaly detection methods. To make these methods able to tackle the spatial and temporal randomness of concurrent noise,
To further distinguish concurrent noise and tackle its spatial and temporal randomness feature, 
we integrate GNN with a novel window-wise graph structure learning technique. This module leverages the information from the potential anomalies identified in the first stage, and avoids the assumptions of stable dependencies among stars or fixed evolution patterns applied in previous methods. By doing this, our model is better equipped to focus on reconstructing concurrent noise, thus enhancing its effectiveness in the context of astronomical observations.

% AERO tackles the first challenge by separately modeling the intra-variate normal patterns and inter-variate concurrent noise in two modules. In the first module, multiple objects are treated as univariate time series to model the inherently temporal normal patterns. After that, anomaly candidates (both true anomalies and concurrent noise) come to the fore through reconstruction error. In the second module, AERO considers all variates together to further model the concurrent noise. In this way, true anomalies are distinguished from concurrent noise because true anomalies remain difficult to fit due to a lack of correlations with other dimensions. This process is shown in Fig.~\ref{fig:flow}(b). To tackle the second challenge, AERO proposes a window-wise graph structure that does not assume a stable association among objects or fixed evolution patterns. It only learns the common noise case in history within every window.

% 贡献
The contributions of our work are summarized as follows:
\begin{itemize}[leftmargin=*]
\item We propose AERO, a novel two-stage time series anomaly detection method tailored for tackling unique characteristics in astronomical observations. To the best of our knowledge, we are the first to systematically identify and address the challenges in astronomical time series anomaly detection.
%astronomical data analysis.
% To the best of our knowledge, we are the first to systematically identify and address the distinct challenges inherent in astronomical observations.
% AERO is the first anomaly detection model that decomposes univariate-normal-relationship and multivariate-noise-relationship. Therefore, it can distinguish true anomalies and concurrent noise better.
\item The proposed AERO employs a Transformer-based encoder-decoder on each variate to select anomaly candidates in alignment with the variate independence property. Then it enhances GNN with a window-wise graph structure learning technique to effectively adapt to the concurrent noise property.
\item We conduct extensive experiments on both synthetic datasets and real-world datasets, the latter obtained from the National Astronomical Observatories of China. Experiments against 11 baseline methods demonstrate that AERO outperforms them in most cases and achieves the highest F1-scores. 
% \footnote{The real datasets will be released upon acceptance.}
% it increases F1-scores by up to 10.48\% and 2.56\% on synthetic and real datasets respectively.
% \item We publicly publish our code and datasets on GitHub for better reproducibility of the results of this paper.
\end{itemize}

\section{Related Work}
\label{s2}
In this section, we present a range of anomaly detection methods relevant to our proposed method. First, we first introduce typical anomaly detection techniques that are applicable to univariate and multivariate time series. Then we concentrate on specialized methods that utilize GNN and Transformer architectures.

\subsection{Time Series Anomaly Detection}
Existing studies can be categorized into two groups of techniques, based on their applicability on either univariate or multivariate time series data.

\textit{\textbf{Univariate techniques}} focus on analyzing a single variate or treating each variate in time series separately, while neglecting their potential correlations. Early studies in this area typically utilize statistical methods. SPOT\cite{SPOT} employs Extreme Value Theory (EVT) to identify outliers of extreme values in streaming univariate time series, bypassing the need for predefined thresholds or assumptions about data distribution. Building on this, FluxEV extends SPOT\cite{FluxEV} by identifying not only extreme values but also abnormal patterns through fluctuation extraction and smoothing processes. In addition, SR\cite{SR} adapts the spectral residual model from computer vision to anomaly detection in industrial services. 
With the advances in deep learning, numerous methods have been developed for anomaly detection based on various models, such as Long Short Term Memory (LSTM)\cite{LSTM-AD,LSTM} and variational auto-encoder (VAE)\cite{Donut,VAE}. Moreover, VAE-LSTM\cite{VAE-LSTM} is a hybrid method that combines VAE for robust local feature extraction over short windows with LSTM for long-term correlation modeling.
However, as discussed in Sec.~\ref{section:1}, these univariate techniques fail to address the issue of concurrent noise encountered in astronomical data scenarios.

\textit{\textbf{Multivariate techniques}} consider a collection of variates in time series as a unified entity. These methods usually model the normal temporal patterns by considering both inter-variate dependencies and variate-specific behaviors, and identify anomalies when data points exhibit large reconstruction errors. Notably,
LSTM-NDT\cite{LSTM-NDT} incorporates a nonparametric anomaly thresholding approach into LSTM for anomaly detection in spacecraft monitoring systems. MSCRED\cite{MSCRED} applies an attention-based convolutional LSTM model for anomaly detection and diagnosis in power plants.  
% Besides LSTM, there are a series of research based on VAE. 
OmniAnomaly\cite{OmniAnomaly} is the first to explicitly account for both temporal dependency and variable stochasticity using VAE. Building upon this idea, InterFusion~\cite{InterFusion}  combines inter-metric correlation and temporal dependency through a hierarchical VAE framework. VQRAEs\cite{VQRAEs} further enables this structure by incorporating bi-directional capability.
In addition, generative adversarial networks (GANs) have been also adapted for anomaly detection in methods like MAD-GAN\cite{MAD-GAN} and DAEMON\cite{DAEMON}. 
% uses an adversarial autoencoder-based architecture. It uses two discriminators to adversarially train an autoencoder to learn the normal pattern. 
TimesNet\cite{TimesNet} is developed as a foundation model applicable to both univariate and multivariate time series anomaly detection tasks. It introduces an approach of transforming time series from 1D to 2D space using Fast Fourier Transform (FFT) and then applies convolution operations to capture temporal and variate dependencies. 
While these methods consider multiple time series as a collective unit, none of them address the issue of concurrent noise. Consequently, they fall short of mitigating false positives caused by this issue.

% OmniAnomaly \cite{OmniAnomaly} is proposed to tackle explicit temporal dependence among stochastic variables to learn robust representations of input data.  
% VQRAEs \cite{VQRAEs} propose an efficient variational quasi-recurrent autoencoder model, which is based on Variational Recurrent Autoencoders(VRAE) and makes it bi-directional.
% VQRAEs \cite{VQRAEs} combines VAE with temporal modeling with specific consideration of achieving good efficiency.

\subsection{GNNs for Time Series Anomaly Detection}
Considering the dependencies among different variates in time series, GNNs~\cite{GNN}, which treat each variate as a node and employ message passing between nodes, have been utilized in multiple time series tasks\cite{surveyGNNTS}, including time series anomaly detection. 
% GNN\cite{GNN} is good at modeling relationships of multivariate, leveraging GNN to explicitly capture relationships between multiple time series is another type of promising approach\cite{surveyGNNTS}. 
GDN\cite{GDN} assumes that variate dependencies (i.e. graph structure) do not change over time, and proposes to capture the static dependencies through embedding learning for each variate. 
% learns an embedding vector for every variate respectively  capture the static topological relation of sensors through sensors' embeddings. 
In line with this assumption, several network structures, such as MTAD-GAT\cite{MTAD-GAT}, Stgat-Mad\cite{Stgat-Mad}, MTGNN\cite{MTGNN}, and RGSL\cite{RGSL}, are developed to learn a static, and globally optimal static graph for multivariate time series.
On the other hand, other methods argue that dependencies among variates and dynamic and evolve over time. These methods, including Event2Graph\cite{Event2Graph}, GraphAD\cite{GraphAD}, BrainNet\cite{BrainNet}, METRO\cite{METRO}, ESG\cite{ESG}, and TSAT\cite{TSAT}, aim to perform dynamic graph structure learning. They achieve this by constructing graphs at every timestamp and employing sequential modeling on the graph structures using RNN~\cite{Event2Graph, METRO, ESG} or Transformer~\cite{TSAT}. Bridging these two assumptions,
SRD\cite{SRD} decomposes variate dependencies into a static graph and is supplemented by dynamic variations unique to individual samples.
While all these methods can selectively learn dependencies among variates, they are not capable of modeling the spatial and temporal randomness characteristic of concurrent noise. 
% none of these methods consider the concurrent noise phenomenon, 
% they don’t model this kind of relationship among different variates. So, these methods cannot be directly used to detect anomalies in catalog streams with co-concurrent noise.

\subsection{Transformer for Time Series Anomaly Detection}
The success of Transformer\cite{Transformer} in sequence modeling has significantly influenced its adoption for time series anomaly detection\cite{surveyTransformer}. 
% Inspired by the success of Transformer\cite{Transformer} in sequence modeling, this architecture has been naturally adopted to handle time series anomaly detection problem\cite{surveyTransformer}. 
Compared to VAE-based methods, Transformer architecture demonstrates its superiority as both an encoder and decoder to account for temporal modeling. Based on this, Transformer variants have been developed in several methods to enhance anomaly detection capabilities. For example, 
% The transformer model inherently cannot explicitly model multivariate relationships. It just treats multivariate at one timestamp as a multidimensional feature. So, many transformer-based models can be used for univariate and multivariate time series.  
TranAD\cite{TranAD} enhances the standard Transformer by integrating self-conditioning and adversarial training processes to improve performance in anomaly detection. 
AnomalyTransformer~\cite{AnomalyTransformer} employs a specialized anomaly attention mechanism to better distinguish normal and abnormal in both univariate and multivariate time series. 
GTA\cite{GTA} augments the input in the Transformer with graph convolutions and hierarchical dilated convolutions to capture variate dependencies and temporal patterns more effectively. Unfortunately, similar to other time series anomaly detection methods, these methods still face challenges in being adequately adapted for astronomical observation scenarios.

% together to form a context encoding block, which describes the topological connection relationships between nodes and dilated convolution captures the long-term temporal dependencies. 
% Then the context encoding block’s outputs are passed into the Transformer-based architecture for encoding and decoding. 
% Similarly, these methods also do not consider concurrent noise, and can not be used to solve our problem directly.

\section{Proposed Method}
\label{section:3}
In this section, we first formulate the problem of anomaly detection in astronomical observations. Next, we introduce an overview of our method, followed by detailed descriptions of each component. Finally, we describe the offline training and online detection process.

\subsection{Problem Statement}\label{sec:problem_statement}
% 图
\begin{figure}[t]
\centerline{\includegraphics[width=0.95\columnwidth]{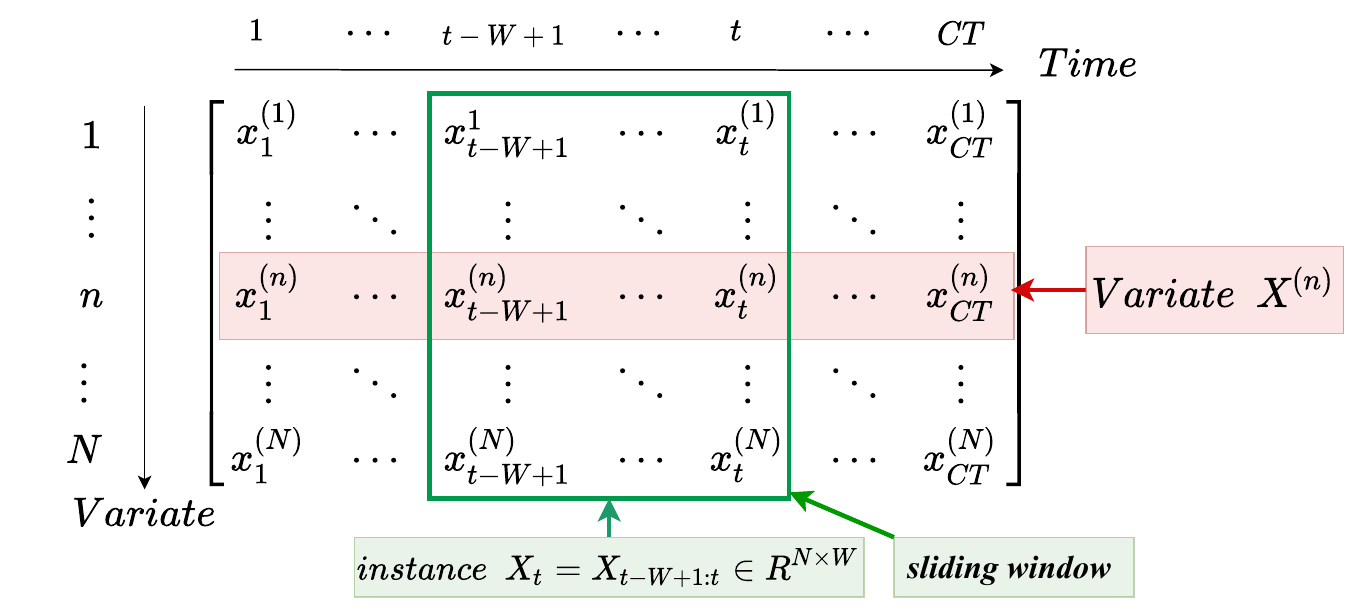}} % Reduce the figure size so that it is slightly narrower than the column. Don't use precise values for figure width. This setup will avoid overfull boxes.
\vspace{-3mm}
\caption{Data format for astronomical observations, which includes $N$ variates (series of magnitudes from $N$ stars) over $CT$ timestamps. 
% Each row $X^{(i)}$ is called an object, while each column $X_t$ is an observation. 
A sliding window of length $W$ is used to partition the complete time series into instances as $X_t \in R^{N \times W}$. }
\vspace{-4mm}
\label{fig:Data formulation}
\end{figure}

Astronomical observation data can be represented as a time series, which consists of $N$ variates (series of magnitudes from $N$ stars) over $CT$ timestamps, denoted as:
\begin{equation}\nonumber
    \mathcal T = \left \{x_1,\cdots,x_{CT} \right \}
\end{equation}
where each datapoint $x_t$ is collected at a specific timestamp $t$ and $x_t = \{ x_t^{(1)},x_t^{(2)},\cdots,x_t^{(N)}\}\in R^{N}$ denotes the magnitudes of $N$ stars at time $t$.

Following the practices in previous studies~\cite{InterFusion,TranAD}, instead of directly utilizing the raw time series $\mathcal T$ as training input, we partition the entire time series into multiple instances by employing a sliding window of length $W$. 
As illustrated in Fig.~\ref{fig:Data formulation}, for a given timestamp $t$, a window of length $W$ generates the instance $X_t$ as follows:
\begin{equation}\nonumber
\begin{aligned}
    X_t = \left \{ x_{t-W+1},\cdots,x_t \right \}\in R^{N\times W}\\
    % where \ x_t=\left \{ x_t^{(1)},x_t^{(2)},\cdots,x_t^{(N)} \right \} \in R^N.
\end{aligned}
\end{equation}
 
In this way, the raw time series $\mathcal T$ is transformed into a collection of instances $\mathcal X = \left \{X_W, X_{W+1}, \cdots, X_{CT} \right \}$ to serve as the data for model training. 
% For each instance $X_t$, we additionally utilize the last $\omega$ datapoints $Y_t =\left\{ x_{t-\omega +1}, \cdots, x_t \right\} \in R^{N \times \omega}$ as the target of reconstruction.  

% We aim to learn a reconstruction function $f(X_t,Y_t)\rightarrow \hat{Y_t}$ capable of effectively representing normal data. By leveraging the reconstruction error, 
Our objective is to determine whether an observation $x_{t}^{(n)}$ for each star at every timestamp is anomalous or not in an unsupervised manner. To achieve this, we aim to learn a predictive function:
\begin{equation}\nonumber
\mathcal{F}(X_t)\rightarrow \mathcal{O}_t
\end{equation}
where $\mathcal{O}_t\in \{0,1\}^{N}$ denotes the binary anomaly labels for N variates at timestamp $t$. By aggregating the results from each instance within the sliding window collection $\mathcal{X}$, we are able to obtain the predictions for the complete time series $\mathcal{T}$.
% Subsequently, employing a threshold $th$, a binary label matrix $O$ is obtained, wherein $o_t^{(i)} \in \left \{0,1 \right \}$, and  $o_t^{(i)} = 1 (s_t^{(i)}\ge th)$ indicates that object $i$ is anomalous at time $t$. This approach facilitates not only the detection of anomalies at the temporal level but also enables their diagnosis at the object level.

\subsection{Overview}

\begin{figure*}[t]
\centerline{\includegraphics[width=1\textwidth]{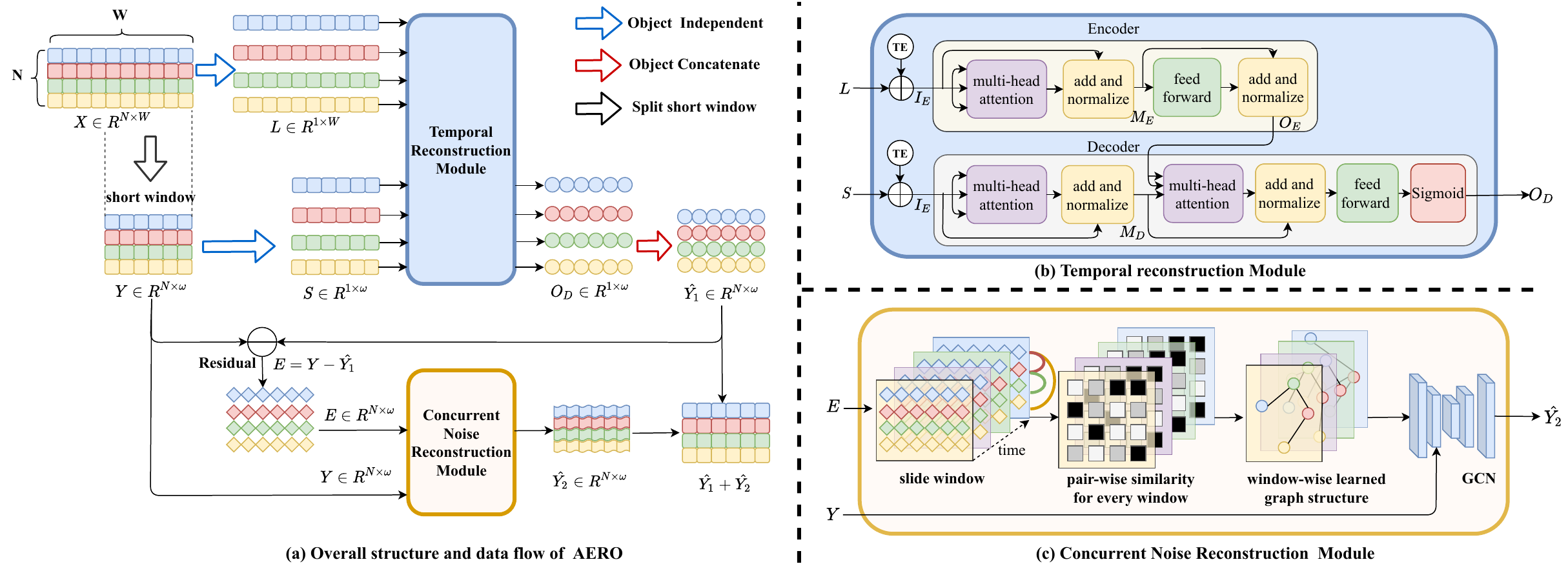}} % Reduce the figure size so that it is slightly narrower than the column.
\caption{An illustration of AERO. (a) The overview framework of AERO. A multivariate time series representing magnitudes series of multiple stars is first divided into univariate time series as the input of the temporal reconstruction module. The initial reconstruction errors from the first module are concatenated as the input of the concurrent noise reconstruction module. (b) Details of temporal reconstruction module. (c) Details of concurrent noise reconstruction module. 
}
\vspace{-3mm}
\label{fig:AERO}
\end{figure*}

% In this section, we will illustrate how our TERO is able to address the two issues in Sec.~\ref{section:1}(i.e.variate independence and concurrent noise with spatial-temporal randomness) towards anomaly detection for astronomical observation data. An overview of TERO is shown in Fig.~\ref{fig:AERO}.
The framework of AERO is illustrated in Fig.~\ref{fig:AERO}. Building on the concept of reconstruction-based anomaly detection, AERO is composed of two modules, namely the temporal reconstruction module and concurrent noise reconstruction module, which are specifically designed to address the two distinct properties (i.e., variate independence and concurrent noise) in astronomical observations. 

% First, AERO aims to learn temporal patterns on each variate from the pespective of a univariate time series and the correlation of concurrent noise among multivariate time series, in turn, using two modules separately. 

The temporal reconstruction module utilizes a Transformer-based encoder-decoder architecture to model the normal temporal patterns of each star. In accordance with the variate independence property, it treats multiple variates as independent univariate time series through a shared network. For a given instance with a window length $W$ from a variate, this module further employs a shorter window with length $\omega$ for reconstruction. This approach ensures that the reconstruction is more focused on the latter parts of the time series while leveraging a longer context to capture temporal patterns better. Such a manner aligns well with the inference stage of anomaly detection at the last timestamp for each instance, as described in Sec.~\ref{sec:problem_statement}. Through this module, anomaly candidates are initially identified based on large reconstruction errors.
% In addition, We use two different kinds of input for encoder and decoder. 
% For more context, we use a complete long window as input of the encoder and then pass the encoded representation to the decoder. When inference online, a newly arrived point is appended to the tail of the slide window so the latter part is more important for its reconstruction. Taking this into consideration, a short window split from the latter part of the complete window is used as input for the decoder. After the first module, anomaly candidates come to the fore with large reconstruction errors. 

The concurrent noise reconstruction module aims to further filter out instances affected by concurrent noise. To achieve this, it models the reconstruction errors from the temporal reconstruction module using a novel graph structure learning technique within GNN. Given the spatial and temporal randomness characteristic of concurrent noise, we devise a simple yet effective window-wise graph structure learning technique, which allows for the generation of a distinct adjacent matrix for every time window. This technique avoids the GNN's tendency to learn stable spatial correlations or predictable temporal patterns in previous methods, which are inconsistent with the randomness in concurrent noise.

% The second module(i.e.Concurrent Noise Reconstruction Module) focuses on modeling the correlations generated by concurrent noise using GNN. To achieve this goal, this module directly analyzes these reconstruction errors outputted by the first module because the patterns of concurrent noise become more distinct and significant making them easier to fit. 

% \textbf{Temporal Reconstruction Module}: The first module is designed for modeling the inherently normal temporal evolution patterns of each object using Transformer. Multiple objects are treated as independent and identically distributed(IID) and share the same network parameter. So we need to divide a multivariate time series representing multiple objects into multiple univariate time series as the input. Meanwhile, we use two types of sliding windows: the long is used for more context as input of the encoder, and the short is a relatively simple reconstruction target as input of the decoder. 
% Second, to tackle the concurrent noise with spatial and temporal random, we propose a simple window-wise graph structure based on GNN in the second module. We calculate a different adjacent matrix for every window and do not analyze the temporal correlations among these adjacent matrices avoiding the assumptions of stable dependencies among objects or fixed evolution patterns applied in previous methods.

Through the integration of these two modules, AERO proficiently reconstructs both normal temporal patterns and concurrent noise. The final anomaly detection results produced by AERO are the combination of both modules. 
% The final anomaly prediction by AERO is a combination of the results from both modules. 
Compared to existing methods, the proposed two-stage framework, tailored for astronomical observations, significantly reduces the number of false positives in practical applications. 
\subsection{Temporal Reconstruction Module}
\label{section:3.3}
% In this module, we employ a modified Transformer architecture to reconstruct normal temporal patterns for every variate. As depicted in Fig.~\ref{fig:AERO}(b), this module consists of five components, including time embedding, input embedding, encoder, decoder, and output layer.

% Transformers originate from natural language and then achieve remarkable performance in many other tasks, such as vision processing tasks\cite{MAE} and time series analysis\cite{surveyTransformer}. 

Transformer has demonstrated its effectiveness in modeling sequential data in various tasks~\cite{BERT, Speech-Transformer}, and therefore has been recently adopted in time series anomaly detection\cite{surveyTransformer}. Its performance surpasses previous reconstruction-based encoder-decoder backbones, such as VAE and RNN\cite{TranAD}. In light of this, we utilize a modified Transformer architecture as a temporal reconstruction module to learn normal temporal patterns through a reconstruction process for each variate. As depicted in Fig.~\ref{fig:AERO}(b), this module consists of five components, including time embedding, input embedding, encoder, decoder, and output layer.

% In light of this, we utilize a Transformer-based architecture as the backbone to learn temporal normal patterns via reconstruction process with the following steps. 
% We now provide details of the temporal reconstruction module. 

\textbf{Time Embedding.} The standard Transformer utilizes positional encoding to integrate order information into a sequence with the latent assumption that intervals between consecutive steps are equal. However, this assumption is not applicable to astronomical observations, which are usually recorded with irregular intervals. 
To consider irregular intervals of observations, 
we propose to adopt an enhanced time encoding technique that incorporates not only the absolute positions in the original trigonometric function but also encodes time intervals as learnable phase shifts\cite{timeembedding1,timeembedding2}.
% some work modifies the conventional sinusoidal absolute positional encoding\cite{timeembedding1,timeembedding2}. 
% Following prior practice, we not only take absolute positions as independent variables in the trigonometric function but also take time intervals as phase shifts. 
Following the practice in~\cite{TranAD}, we sum sin and cos terms as the final time embedding function, which produces improved performance. Then the $j$-th dimension of time embedding $TE_t$ at timestamp $t$ is defined as:
\begin{equation}
\begin{aligned}
    TE_t^j&=sin(f^j\times pos_t + \alpha _j \times\Delta_t) \\
        &+cos(f^j\times pos_t + \alpha _j \times\Delta_t)
\end{aligned}
\end{equation}
where $f^j$ represents the pre-defined angle frequency, calculated as $f^j=(1/10000)^{j/d_m}, j\in[0,d_m]$, $d_m$ is the dimensions of hidden states in Transformer, $pos_t$ is the absolute position of timestamp $t$, $\Delta_t$ is the time interval between the current timestamp and the previous one, and $\alpha_j$ is a learnable parameter.

\textbf{Input Embedding.} For the reconstruction process, we generate two types of model inputs, $X_t \in R^{N\times W}$ derived from long sliding windows, and its subsequence $Y_t$ in the latter part with a short window length of $\omega$ $(\omega < W)$, written as follow: 
\begin{equation}\label{Yt}
    Y_t= \left \{ x_{t-\omega +1},\cdots,x_t \right \} \in R^{N\times\omega}
\end{equation}

The rationale behind this strategy is aligned with the inference stage of time series anomaly detection, where anomaly scores are progressively determined for the last timestamps using sliding windows. Consequently, our primary focus is on the reconstruction of the latter parts of a time series (i.e., $Y_t$), while still leveraging a longer context (i.e., $X_t$) for the effective modeling of temporal patterns. 

% In order to provide a long context, the encoder takes a long window as input and then passes the encoded representation to the decoder for better reconstruction. For a more accurate reconstruction, the decoder takes a short window as input. Besides, multiple objects are viewed as IID univariate sequences in this module. 

% Formally, a short window of length $\omega$ is split from the latter part of $X_t \in R^{N\times W}$ as:
% \begin{equation}
%     Y_t= \left \{ x_{t-\omega +1},\cdots,x_t \right \} \in R^{N\times\omega}
% \end{equation}
% where $\omega < W$. 
Incorporating the prior knowledge of variate independence in astronomical observations, we model $X_t$ and $Y_t$  separately as $L_t^{(n)} \in R^{1 \times W}$ and $S_t^{(n)} \in R^{1\times \omega}$ as follows: 
\begin{equation}
\begin{aligned}
     L_t^{(n)} = X_t^{(n)} = \left \{ x_{t-W+1}^{(n)},\dots ,x_t^{(n)} \right \},  \\
     S_t^{(n)} = Y_t^{(n)} = \left \{ x_{t-\omega+1}^{(n)},\dots ,x_t^{(n)} \right \}.  \\
\end{aligned}
\end{equation}

Then, $L_t^{(n)}$ and $S_t^{(n)}$ are projected into an $d_m$ dimensional embedding by a linear projection. We add the time embedding $TE_t$ to the projected time series embeddings as the final input embedding $IE_t^{(n)}$ and $ID_t^{(n)}$ as follows:
\begin{equation}
\begin{aligned}
     IE_t^{(n)} = W_E \times L_t^{(n)} + TE_t, \\
     ID_t^{(n)} = W_D \times S_t^{(n)} + TE_t.  \\
\end{aligned}
\end{equation}

% We add the time embedding $TE_t$ to the time series embeddings $IE_t^{(n)}$ and $ID_t^{(n)}$  as the final input embedding for the subsequent components:
% \begin{equation}
% \begin{aligned}
%     I_{Et}^{(n)} = IE_t^{(n)} + TE_t, \\
%     I_{Dt}^{(n)} = ID_t^{(n)} + TE_t. \\
% \end{aligned}
% \end{equation}

To maintain clarity while simplifying the discussion, we omit superscripts denoting variates and subscripts denoting timestamps. Therefore, we substitute $I_E$ and $I_D$ for $IE_t^{(n)}$ and $ID_t^{(n)}$ respectively in our subsequent explanations.
% \begin{small}
% \begin{align}
%     \label{eq1} L_t^{(n)} = X_{t-W_L:t-1}^{(n)} = \left [ x_{t-W_L}^{(n)},x_{t-W_L+1}^{(n)},\dots ,x_{t-1}^{(n)} \right ]  \\
%     \label{eq2} S_t^{(n)} = X_{t-W_S:t-1}^{(n)} = \left [ x_{t-W_S}^{(n)},x_{t-W_S+1}^{(n)},\dots ,x_{t-1}^{(n)} \right ]  \\
%     I_t^{(n)} = W_E \times L_t^{(n)}, or, I_t = W_D \times S_t^{(n)}  
% \end{align}
% \end{small}%

\textbf{Encoder.} The encoder produces the representations based on the time series of long window size. The representations are generated based on the self-attention mechanisms applied in Transformer architecture. Specifically,
% The representations, conditioned on longer context information, are then passed to the decoder helping to reconstruct the short window better. 
the self-attention mechanisms perform the following operations:
\begin{equation}
    Attention(Q,K,V)=softmax(\frac{QK^T}{\sqrt{d_m}})V.
\end{equation}
where $Q$, $K$, and $V$ represent the query, key, and value matrix respectively derived from a linear projection on the input embeddings.

In our work, we adopt multi-head self-attention to model the embeddings of time series. Specifically, the input embeddings are projected into $h$ sets of different queries, keys, and values to perform self-attention mechanism, which has been shown to achieve better performance. Given the input representations $I_E$, the output representations of multi-head self-attention are produced as follows:
\begin{equation}
\begin{aligned}
    MHA(Q,K,V)&=Concat(H_1,...,H_h)\times W_O \\
    H_i &= Attention(Q_i,K_i,V_i), 
    % {\rm where}\ Attention(Q_i,K_i,V_i)=softmax(\frac{Q_iK_i^T}{\sqrt{m}})V_i.
\end{aligned}
\end{equation}
where $Q_i$, $K_i$, and $V_i$ for $i\in\{1,\dots,h\}$ are obtained by passing input $I_E$ through projection matrices $W^{i}_{Q}$, $W^{i}_{K}$, $W^{i}_{V}$$\in\mathbb{R}^{d_m \times d_m/h}$ for each head $h$, $W_O$ is learnable parameter matrix.

We follow standard Transformer architecture to combine the residual connection and layer normalization with multi-head self-attention, which can be expressed as follows:
% 多行公式单编号
\begin{equation}
    \begin{aligned}
        M_E &= LayerNorm(I_E + MHA(I_E, I_E, I_E)),\\
        O_E &= LayerNorm(M_E + FFN(M_E)).  \\
    \end{aligned}
\end{equation}

where $LayerNorm$ denotes layer normalization operation, $FFN$ represents the feed-forward neural networks, and $O_E \in R^{d_{m}\times W}$ is the final output representations derived from the encoder. 

% where $MHA$ represents multi-head attention, and $FFN$ represents the feed-forward neural networks. 

% Multi-head attention allows the model to jointly attend to information from different representation subspaces at different positions. We apply MHA by first passing input $I_E$ through $h$ (number of heads) feed-forward layers to get $Q_i$,$K_i$, and $V_i$ for $n\in\{1,\dots,h\}$, where $h = \left \lfloor D_{model}/2 \right \rfloor $ and then applying scaled-dot product attention. 

% \begin{small}
% \begin{align}
%     &MHA(Q,K,V)=Concat(H_1,...,H_h) \\
%     {\rm where}\ &H_i=Attention(Q_i,K_i,V_i) \\
%     {\rm where}\ &Attention(Q_i,K_i,V_i)=softmax(\frac{Q_iK_i^T}{\sqrt{m}})V_i
% \end{align}
% \end{small}

% Here, the softmax forms the convex combination weights for $V_i$, compressing the matrix $V_i$ into a small representative embedding.

\textbf{Decoder.} The objective of the decoder is to reconstruct the time series with a shorter window length. To achieve this, the decoder takes the embeddings $I_D$, and the output representations $O_E$ of the encoder as contextual information to reconstruct the time series for each variate. Specifically, the decoder performs the following operations:
\begin{equation}
    \begin{aligned}
        M_D&=LayerNorm(I_D + MHA(I_D, I_D, I_D)), \\
        O_D^{'}&=LayerNorm(M_D + MHA(M_D, O_E, O_E))
    \end{aligned}
\end{equation}

% The encoding of the long window $O_E$ from the encoder is used as value and keys for the decoder's attention operation using the encoded short window as the query matrix. In other words, the encoder has generated attention weights using the long context window for the decoder to capture temporal trends within the input sequence.
The representations from the encoder are used as values and keys for the queries generated by shorter windows to capture temporal patterns within the long context.

Finally, feedforward neural networks and sigmoid activation are employed to generate the normalized predictions of a variate: 
\begin{equation}
        O_D = Sigmoid(FFN(O_D^{'})).\\
\end{equation}
% 多行公式多编号
% https://blog.csdn.net/Strive_For_Future/article/details/118609968

\textbf{Output Layer.} 
% Since the temporal reconstruction module processes every object independently and can not enter the next module directly, we need to concatenate multiple $O_D$ of every object together to form an entity $\hat{Y_1}\in R^{N \times \omega}$ as a multivariate time series.
In the output layer, we concatenate the result $O_D\in R^{1\times \omega}$ from each variate to produce a matrix, denoted as $\hat{Y_1}\in R^{N \times \omega}$, as a reconstructed multivariate time series:
% Formally, we perform the following operation:
\begin{equation}
    \hat{Y_1} = Concat(O_D^{(1)},O_D^{(2)},\dots,O_D^{(N)}).
\end{equation}

Subsequently, we calculate the initial reconstruction errors $E\in R^{N \times \omega} $ as follows:
\begin{equation}
    E = Y-\hat{Y_1}.
\end{equation}

By doing this, anomaly candidates are identified by relatively large errors, whereas normal patterns are characterized by smaller errors. 
% To sum up, this module captures each object's temporal evolution pattern independently. As a result, normal data points are reconstructed accurately, while anomaly candidates(both true anomalies and concurrent noise) are difficult to fit in this module and exhibit larger reconstruction errors. In this way, we first distinguish between normal pattern and anomaly candidates.

\subsection{Concurrent Noise Reconstruction}
The temporal reconstruction module is effective at detecting anomalies on a variate-wise basis. However, there is a high likelihood of concurrent noise being mistakenly classified as anomalies if the correlations among variates are not taken into account. In this module, we aim to refine this issue by filtering out concurrent noise from the identified anomaly candidates. Since concurrent noise tends to manifest large errors simultaneously across multiple variates, it can typically be distinguished through the modeling of variate correlations. As illustrated in Fig.~\ref{fig:AERO}(c), we further refine the anomaly detection process by reconstructing the errors obtained from the temporal reconstruction module. This is achieved by applying GNN enhanced with a window-wise graph structure learning technique detailed as follows.   
% mechanism
% After the univariate temporal reconstruction module, anomaly candidates still exhibit larger reconstruction. Excluding the influence of various patterns in the origin series, directly analyzing the reconstruction errors $E$ from the first module is more distinct and clear for the second module to model the correlation of concurrent noise. Among anomaly candidates, multiple concurrent noise has correlations and can be easily fitted by GNN, while true anomalies remain difficult to fit since they lack correlations. In this way, we can further distinguish concurrent noise and true anomalies by reconstruction error in this module.
% We now provide details of the concurrent noise reconstruction module. Module structure as shown in Figure \ref{fig:AERO} (c).

\textbf{Window-wise Graph Structure Learning.} 
Concurrent noise exhibits characteristics of spatial and temporal randomness, implying that it can appear on an unpredictable number of stars at any time period. Such characteristics render the application of existing static or dynamic GNN methods unsuitable. Static GNN methods rely on a static graph structure to represent stable dependencies among variates, while dynamic GNN methods operate under the assumption of predictable evolving patterns in variate correlations. To overcome the limitations, we propose a novel approach that involves constructing a graph structure specific to each sliding window. Specifically, we leverage the errors generated by the temporal reconstruction module, denoted as $E_t\in R^{N\times\omega}$, as the embedding for the sliding window at timestamp $t$. Then we compute the similarity between variates $m$ and $n$ in $E_t$ as follows:
% This is different from classical static GNN and dynamic GNN since they are not appropriate. On the one hand, since concurrent noise has spatial randomness, there is no stable and invariant association among multiple objects suitable for classical static GNN. On the other hand, concurrent noise has temporal randomness lacking regular evolution patterns so we can not utilize temporal correlation among graph sequences in dynamic GNN. 
% In this case, we generate different adjacency matrices for every window to model random concurrent noise. Specifically, we treat object $n$ as a node and take corresponding error window $E_t^{(n)}\in R^{1\times\omega}$ as node embedding, then compute similarity between node $m$ and $n$ as:
\begin{equation}
    sim_t^{mn} = \frac{(E_t^{(m)})^T E_t^{(n)}}{\left \| E_t^{(m)} \right \| \cdot \left \| E_t^{(n)} \right \|} 
    % ,for \ m,n\in \{1,2,\dots, N\}
\end{equation}
% E_t^n \in R^{s\times N}

Based on this, the graph structure corresponding to the window at timestamp $t$ is determined by an adjacency matrix $A_t$, which indicates the pairwise similarity between variates: 
\begin{equation}
    A_t^{mn}=sim_t^{mn}
\end{equation}

% \begin{small}
% \begin{align}
%     A_t^{ij} &= sim_t^{ij}
%     % A_t &= \begin{bmatrix}  &sim_t^{11}  &\cdots   &sim_t^{1i}  &\cdots & sim_t^{1j} &\cdots & sim_t^{1N} \\  &\vdots   &\ddots  &\vdots &\ddots  &\vdots &\ddots  &\vdots\\  &sim_t^{i1}  &\cdots  &sim_t^{ii}  &\cdots &sim_t^{ij} &\cdots &sim_t^{iN}\\  &\vdots   &\ddots  &\vdots &\ddots  &\vdots &\ddots  &\vdots\\  &sim_t^{N1}  &\cdots  &sim_t^{Ni}  &\cdots &sim_t^{Nj} &\cdots &sim_t^{NN}\\\end{bmatrix}
% \end{align}
% \end{small}%
    
\textbf{Reconstruction via GCN.} 
After deriving the graph structure for each sliding window, variates exhibiting similar error patterns are assigned large weights within the graph. 
% Then we apply a GNN to perform message passing among variates
% To capture the relationships of concurrent noise, we use a spectral graph convolutional network(GCN) as a feature extractor. 
As illustrated in Fig.~\ref{fig:catalog}, variates affected by concurrent noise demonstrate simultaneous fluctuations, whereas true anomalies usually exhibit distinct temporal deviations. This principle is critical in distinguishing concurrent noise from true anomalies. In other words, if a variate is influenced by concurrent noise within a given window, it can be effectively reconstructed using the error patterns of other similarly affected variates. This principle, however, does not apply to true anomalies. Leveraging this insight, we employ a GNN to perform message passing among variates for concurrent noise reconstruction. The GNN is implemented as follows: 
% On the contrary, if node $n$ contains a true anomaly, it is not related to other nodes, so other nodes can not reconstruct it. 
% As a result, concurrent noise and true anomaly can be distinguished by reconstruction error.
% Specifically, we only use embedding of other nodes to represent node $n$ instead of using the node's own embedding. This design can avoid true anomaly being well reconstructed according to its own information. 
% Based on the obtained adjacency matrix, we fuse a node's information with its first-order neighbors using the Chebyshev polynomial as the filter like \cite{GCN}:
\begin{equation}
\begin{aligned}
    % Y_2 =tanh((\tilde{D}^{-\frac{1}{2}} \tilde{A} \tilde{D} ^{-\frac{1}{2}} Y_t)\theta +\beta)
    \hat Y_2 =\sigma((\tilde{D}^{-1} \tilde{A} Y_t)W_{\theta} + b_{\theta})
\end{aligned}
\end{equation}
%     Z_t &= tanh((I+D^{-\frac{1}{2} } \widetilde{A}_t D^{-\frac{1}{2} })S^n \theta +\beta ) \\
% {\rm where} \ \widetilde{A}_t &= A_t - I
where $Y_t$ is the input with the short window size in Eq.~(\ref{Yt}), $\tilde{A}= A-I$, $\tilde{D}$ is the degree matrix, and $\tilde{D}^{mm}= {\textstyle \sum_{n}} \tilde{A}^{mn}$, $\sigma$ is the activation function, and $W_{\theta}$ and $b_{\theta}$ are learnable parameters. It is important to note that self-loops are intentionally removed in message passing to exclude the information of the target node itself. This design avoids the situations in which true anomalies are well reconstructed based on their own information. 

% \textbf{Output.} From the above feature extractor, we obtain representations for all windows, namely $\{Z_1\,\dots, Z_CT \}$. We stack them as the final output of the second module:
% \begin{equation}
%     \hat{Y_2} = Concat(Z_1,Z_2,\dots,Z_CT)
% \end{equation}

\subsection{Offline Two-Stage Model Training}
% In order to separately focus on modeling normal data and concurrent noise, 
To separately focus on modeling normal temporal patterns and concurrent noise, the training of two modules in AERO is sequentially arranged in two stages. 
The overall training process is described in Algorithm \ref{alg:train}.

\textbf{Stage 1: Normal Temporal Pattern Reconstruction.}
In the first stage, we train the temporal reconstruction module 
% while the second module is frozen, aiming to model the temporal pattern to the maximum extent while excluding the interference of the second module. 
After this stage, concurrent noise and true anomaly become prominent by large reconstruction errors. 
% shielding the interference of various patterns of origin input. 
When the loss of the first module does not decrease over several patient epochs, we stop training the temporal reconstruction module and enter the second stage to handle concurrent noise. The loss function of the first stage is:
\begin{equation}
    loss_{rec} = Y-\hat{Y_1}
\end{equation}
% where $Y - \hat{Y_1}$ is reconstruction error from the temporal reconstruction module.

\textbf{Stage 2: Concurrent Noise Reconstruction.}
In the second stage, we train the concurrent noise reconstruction module while freezing the parameters of the first module to maintain stable training. The loss function is defined as: 
% aiming to mine the correlation of concurrent noise especially based on the reconstruction error of the first module. 
% After further fitting concurrent noise, the final output is the sum of the outputs of two modules. So the loss function of the second stage is:
\begin{equation}
    loss_{noise} = Y-\hat{Y_1}-\hat{Y_2}
\end{equation}

% 在线测试
\subsection{Online Detection and Diagnosis}
After the model training, we can perform online anomaly detection, which is summarized in Algorithm \ref{alg:test}. 
% At train time, we only look back at the data within a window of length $W$ until the current timestamp. Hence 
During the inference phase, AERO operates in an online mode by following the patterns adopted in the training stage. This involves maintaining a sliding window with a stride 1. As new observations arrive, they are appended to the preceding window. Then the model determines whether the data points are anomalous or not based on the anomaly scores, which are defined as the combination of reconstruction errors in two modules: 
\begin{equation}
    s_t = \mathcal{S}(Y-\hat{Y_1}-\hat{Y_2}).
\end{equation}

where $\mathcal{S()}$ represents the indexing function that selects the last timestamp to produce $s_t \in R^{N}$. Then for each variate, a higher anomaly score $s_t^{(n)} \in R$ indicates that the corresponding input $x_t^{(n)}$ is hard to reconstruct, and thus more likely to be an anomaly. 

Based on the anomaly scores for $x_t^{(n)}$, we can derive point-wise anomaly labels for each variate. Specifically, if $s_t^{(n)}$ is larger than a threshold, the corresponding $x_t^{(n)}$ is marked as an anomaly. We utilize the widely adopted Peak Over Threshold (POT) method in previous studies~\cite{SPOT, OmniAnomaly, TranAD, SR} to automatically determine the threshold. The final anomaly label is derived by:

% On the contrary, the smaller the score, the less likely the observation is anomalous. 

% Based on the point-wise anomaly score, we can get a point-wise anomaly label. Formally, if $s_t^{(n)}$ is larger than an anomaly threshold, $x_t^{(n)}$ is marked as anomalous(1). We adopt the Peak Over Threshold(POT) method to choose the threshold of anomaly score automatically. The final predictive label is determined by:
\begin{equation}
    \mathcal O_t^{(n)} = \mathbbm{1}(s_t^{(n)}\ge POT(s))
\end{equation}
where $s$ utilized in the POT method is the collection of all anomaly scores obtained from the training instances, and  $s_t^{(n)}$ denotes the anomaly score for variate $n$ at the current time $t$.
% \begin{equation}
% \begin{aligned}
%     s =\left \{&s_1^{(1)}, s_2^{(1)},\cdots,s_{CT}^{(1)},\\
%     &s_1^{(n)},s_2^{(n)},\cdots,s_{CT}^{(n)}, \\
%     &s_1^{(N)}, s_2^{(N)},\cdots,s_{CT}^{(N)}\right \}
% \end{aligned}
% \end{equation}

% POT is a commonly used method for threshold selection. This method is the second theorem in Extreme Value Theory(EVT). The basic idea is to fit the tail portion of a probability distribution by a generalized Pareto distribution (GPD) with parameters.  
% SPOT\cite{SPOT} first uses it to detect anomalies in a stream and OmniAnomaly \cite{OmniAnomaly} first uses it to determine the threshold based on SPOT. So does TranAD\cite{TranAD} and SR\cite{SR}. 

% Based on the point-wise anomaly label, we can diagnose the anomaly not only at the variate level but also at the time level directly. While for many other models, an extra time-consuming procedure is needed for diagnosis\cite{DAEMONWSDM}.

\begin{algorithm}[tb]
\caption{Model Training Process}
\label{alg:train}
% \small
\footnotesize
% \scriptsize
% \tiny
\textbf{Input}: $X$ and $Y$ split from $\mathcal T$ using sliding windows\\
\textbf{Require}: Temporal Reconstruction Module $M_1$,\par
 \quad Concurrent Noise Reconstruction Module $M_2$, \par
 \quad $epoch_1$ and $epoch_2$ decided by early stop mechanism.\\
\textbf{Output}: Trained $M_1$ and $M_2$
\begin{algorithmic}[1] %[1] enables line numbers

\FOR{$i=0$ to $epoch_1$} 
    % \STATE $stage=1$.
    \STATE $\hat{Y_1} \leftarrow M_1(X, Y)$
    \STATE $loss_1 = Y - \hat{Y_1}$
    \STATE Update parameter of $M_1$ by $loss_1$.
\ENDFOR
\FOR{$i=epoch_1$ to $epoch_2$} 
    % \STATE $stage=2$.
    \STATE $\hat{Y_2} \leftarrow M_2(Y-\hat{Y_1},Y)$
    \STATE $loss_2 = Y - \hat{Y_1} - \hat{Y_2}$
    \STATE Update parameter of $M_2$ by $loss_2$.
\ENDFOR
\end{algorithmic}
\end{algorithm}

\begin{algorithm}[tb]
\caption{Online Detection Process}
\label{alg:test}
% \small
\footnotesize
% \scriptsize
% \tiny
\textbf{Input}: $X$ and $Y$ split from test dataset using sliding windows\\
\textbf{Require}: Trained $M_1$ and $M_2$. \\
\textbf{Output}: Predicted label matrix $\mathcal O$.\par
\begin{algorithmic}[1] %[1] enables line numbers

\FOR{$t$ at each timestamp}
    \STATE $\hat{Y_1} \leftarrow M_1(X,Y)$
    \STATE $\hat{Y_2} \leftarrow M_2(Y-\hat{Y_1},Y)$
    \STATE $ s_t = \mathcal{S}(Y-\hat{Y_1}-\hat{Y_2})$
    \FOR{each variate $n$}
    \STATE $\mathcal O_t^{(n)} = \mathbbm{1}(s_t^{(n)}\ge POT(s))$
    \ENDFOR
\ENDFOR
\end{algorithmic}
\end{algorithm}

\vspace{-2.5mm}

\section{Experiments}
\label{section:4}
In this section, we conduct experiments to answer the following research questions:
\begin{itemize}[leftmargin=*]
    \item \textbf{RQ1:} Can our method outperform baselines for anomaly detection in astronomical observations by improving precision while guaranteeing recall? 
% Does our method outperform baselines and successfully decrease the number of false positives for anomaly detection in astronomical observations?
    \item \textbf{RQ2:} How do different components of our method benefit the performance?
    \item \textbf{RQ3:} How does our method perform in terms of efficiency and scalability?
    \item \textbf{RQ4:} Can the learned graph structure in our method effectively represent the concurrent noise? How does each step of our method contribute to the anomaly scores?
    \item \textbf{RQ5:} How do the hyperparameters and configurations influence our method performance? 
    % \item \textbf{RQ5:} How does the window size influence the model performance? 
\end{itemize}

\subsection{Datasets}
We utilize six datasets, including three synthetic datasets and three real-world datasets, to evaluate the performance of compared methods.
\begin{table}[t]
\footnotesize
\caption{Dataset Statistics. }
\vspace{-5mm}
\begin{center}
\setlength{\tabcolsep}{0.5mm}{
% \resizebox{1\columnwidth}{!}{
% \resizebox{\textwidth}{6mm}{
\begin{tabular}{@{}lcccccccc@{}}
\toprule
Dataset            & \#train & \#test & \makecell{\#vari-\\ates} & \makecell{\textbf{A}nomaly\\(\%)}    & \makecell{\textbf{N}oise\\(\%)} &\textbf{A/N}     & \makecell{\#Anomaly\\ Segments} & \makecell{\#Noise\\ variates}  \\ \midrule
SyntheticMiddle     & 4000    & 4000   & 24 & 0.180    & 1.719  & 0.105   & 5    &  17/24    \\
SyntheticHigh       & 4000    & 4000   & 24 & 0.359    & 1.719  & 0.209   & 10   &  17/24    \\
SyntheticLow        & 4000    & 4000   & 24 & 0.180    & 3.438  & 0.052   & 5    &  17/24    \\
AstrosetsMiddle     & 5540    & 5387   & 54 & 0.153    & 4.173  & 0.037   & 2    &  54/54    \\
% (024\_G0014-03250425)
AstrosetsHigh       & 8000    & 6117   & 38 & 0.117    & 2.405  & 0.049   & 2    &  38/38    \\
% (044\_G0014-02820255_2) 
AstrosetsLow        & 6255    & 2950   & 40 & 0.190    & 8.419  & 0.023   & 6    &  40/40    \\
% (032\_22790425-G0014) 
\bottomrule 
\multicolumn{9}{l}{*Anomaly(\%) represents the proportion of anomalous data points.}\\
\multicolumn{9}{l}{*Noise(\%) is the proportion of data points affected by concurrent noise.}\\
\multicolumn{9}{l}{\makecell[l]{*A/N denotes the anomaly-to-noise ratio which measures the ratio of true \\anomalies in the potential candidates.}}\\
\multicolumn{9}{l}{*\#Noise variates is the number of variates affected by concurrent noise.}
\end{tabular}
}
\label{table:Dataset}
\end{center}
\vspace{-5.5mm}
\end{table}

\textbf{Synthetic Datasets.} Astronomical observations are unique in exhibiting characteristics of variate independence and concurrent noise, which usually do not exist in other data domains. To showcase the effectiveness of our method, we generate three synthetic datasets in which these properties are injected into time series. 
% In a nutshell, we inject three kinds of true anomalies into two kinds of basic signals. 
% Specifically, we first generate basic signals randomly obey $X\sim N(0,0.2^2)$ Gaussian distribution simulating non-variable star or sinusoidal function distribution added Gaussian noise simulating variable star. The sinusoidal function is:
The construction of these datasets begins with the generation of basic signals. These signals either conform to a Gaussian distribution, $X\sim N(0,0.2^2)$, to simulate the behavior of non-variable stars, 
or obey a sinusoidal function with added Gaussian noise, thereby imitating the behavior of variable stars. The employed sinusoidal function is as follows:
\begin{equation}\nonumber
    f(t, T) = 2*sin(\frac{2*\pi}{T}*pos_t)
\end{equation}
where $pos_t$ is the absolute position of the current timestamp and $T$ denotes the cycle value sampled from a range between 100 and 300 simulating various variable stars.

Then three types of concurrent noise are injected into the basic signals. The first type is data drift, which is simulated by increasing or decreasing the mean value in the basic signals. The second type represents the process of darkening followed by recovery, which is induced by occlusions such as cloud cover. This effect is simulated by adding half a period of trigonometric function to the basic signals at specific timestamps.
% This effect is simulated by adding trigonometric functions to the basic signals for a period of time.
% This is simulated by injecting signals obeying trigonometric functions to a few basic signals at the same timestamps. 
The third type represents the brightening effect, caused by the sunrise in the morning. This effect is simulated by adding exponential functions to the basic signals for a period of time.

\begin{figure}[ht]
\centerline{\includegraphics[width=0.5\textwidth]{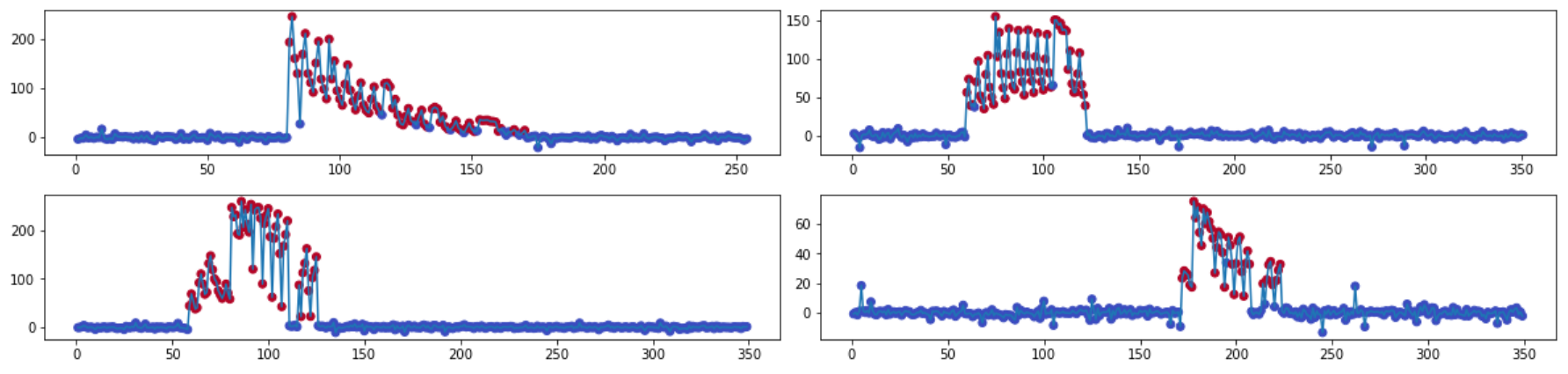}} 
\vspace{-2mm}
\caption{Examples of Injected True Anomalies.}
\vspace{-3mm}
\label{fig:simulation}
\end{figure}

As for the injected true anomalies, we use two categories in astronomical classification datasets from kaggle\footnote{https://www.kaggle.com/competitions/PLAsTiCC-2018/data} and flare function in\cite{flare}. Examples of true anomalies are shown in Fig.~\ref{fig:simulation}.
We randomly inject the true anomalies to the basic signals across various variates to create the SyntheticMiddle dataset with a moderate anomaly-to-noise ratio (A/N). Similarly, we generate additional synthetic datasets with varying A/N ratios. Specifically, we either double the number of anomalous data points, or the amount of concurrent noise to create the SyntheticHigh and SyntheticLow datasets. The statistics of the datasets are summarized in Table \ref{table:Dataset}.

\textbf{Real-world Datasets.} Given the absence of public datasets for anomaly detection in astronomical observations, we have curated three real-world datasets, referred to as Astrosets.
% \footnote{The datasets will be released upon paper acceptance.}. 
These datasets are processed from the astronomical observations conducted by the Ground-based Wide Angle Cameras (GWAC)\cite{LGW} at the National Astronomical Observatories of China. The three datasets are selected to contain a broad range of statistical characteristics, as presented in Table \ref{table:Dataset}. 
% Notably, Anomaly(\%) represents the anomaly points number to total length ratio. 
% similarly, Noise(\%) represents the concurrent noise points number to total length ratio. 
% In order to reflect the relationship between the number of anomalies and the number of concurrent noises more accurately and directly, we calculate A/N as key information and name three datasets according to this feature. 
Moreover, different from other anomaly detection datasets, anomalies in real astronomical observations are relatively rare. In light of this, we specifically annotate the number of segments where anomalies occur to illustrate the datasets. 
% Finally, column \#Noise Objects means a number of objects that occur concurrent noise.
% numbers of objects, different amounts of concurrent noise, and different true anomalies (rare scientific phenomena, such as stellar flares). 
% According to different ratios of anomaly and concurrent noise, we name them AstrosetsMiddle, AstrosetsHigh, and AstrosetsLow.

\subsection{Baselines and Experimental Setup}
We compare the performance of our proposed AERO\footnote{Codes and datasets are available at: https://github.com/XinliHao/AERO} with ten methods for time series anomaly detection, including five univariate methods (i.e., Template Matching, SR, SPOT, FluxEV, and Donut) and six multivariate methods (i.e., OmniAnomaly, AnomalyTransformer, TranAD, GDN, ESG, and TimesNet). Details of these baselines are provided as follows:

\begin{itemize}[leftmargin=*]
    \item Template Matching (TM)\cite{SciDetector}: it is a supervised method for celestial event discovery in astronomy. It employs pre-defined event templates to match newly arrived data. We treat the templates as anomalies in the experiments.
    \item \textbf{SR} \cite{SR}: it
% \footnote{https://paperswithcode.com/paper/time-series-anomaly-detection-service-at\#code}
is a univariate time series anomaly detection method based on Spectral Residual~\cite{saliency}. Since SR does not require training, we directly apply it in online detection. 
% Other settings are the same as the original paper.

    \item \textbf{SPOT}\cite{SPOT}:
% \footnote{https://github.com/Amossys-team/SPOT} 
it applies Extreme Value Theory to detect anomalies in univariate streaming data. 
% We set the initial threshold $level=0.99$ and desired probability $q=10^{-3}$ for every object.

    \item \textbf{FluxEV}\cite{FluxEV}: it augments SPOT with the capability of identifying not only extreme values but also a range of abnormal patterns. % We set the same initial threshold $t$ and desired probability $q$ as SPOT. We follow the other settings in the original paper.

    \item  \textbf{Donut}\cite{Donut}: it utilizes a variational auto-encoder (VAE) as the backbone to serve as a reconstruction-based method for univariate time series anomaly detection. 
    % We set latent dimension $K=5$ and other settings are the same as in the original paper.

    \item  \textbf{OmniAnomaly (OA)}\cite{OmniAnomaly}: it employs VAE to explicitly model the dependencies among variates in a stochastic manner for multivariate time series anomaly detection. 
% The length of the input sequence is set to 200 following our AERO. Other settings are the same as the original paper.

    \item \textbf{AnomalyTransformer (AT)}\cite{AnomalyTransformer}: it adapts the Transformer as a reconstruction-based anomaly detection model by incorporating an anomaly-attention mechanism and an association discrepancy analysis method. We transform the univariate anomaly score into a multivariate result for comparison. 
    % by multiplying it and the multivariate reconstruction error together.
    % We multiply the joint criterion and reconstruction error together to form multivariate results for a fair comparison. 
% The length of the input sequence is set to 200 following our AERO. Other settings are the same as the original paper.

    \item \textbf{TranAD}\cite{TranAD}: it incorporates score-based self-conditioning adversarial training into Transformer encoder-decoder architecture for multivariate time series anomaly detection. 
    % The window size is 10 following the original paper.
% The length of the input sequence is set to 200 following our AERO. Other settings are the same as the original paper.

    \item \textbf{GDN}\cite{GDN}: it is a GNN-based method that explicitly learns a global static graph structure to indicate the correlations among variates through embedding techniques. 
    % We use embedding vectors with a length of 64. The number of neighbor nodes $k=4$ for Synthetic datasets and 9,6,7 for three real-world datasets separately which are about 1/6 of the total variates.

    \item  \textbf{ESG}\cite{ESG}: it is a foresting model based on a dynamic GNN that learns a dynamic graph structure for variates. We adapt it for anomaly detection by employing single-step prediction errors. 
% The length of the input sequence is set to 200 following our AERO. Other settings are the same as the original paper.

    \item \textbf{TimesNet}\cite{TimesNet}: it is the state-of-the-art foundation model for various tasks in time series analysis, including anomaly detection tasks. It introduces a novel approach that applies convolutions to time series by transforming them into 2D space. 

\end{itemize}

For all the baselines in our experiments, we use the implementations provided by the authors or from well-established repositories on GitHub. 
% \footnote{https://paperswithcode.com/paper/time-series-anomaly-detection-service-at\#code} 
% \footnote{https://github.com/Amossys-team/SPOT}. 
We set the length of the input sequence to be consistent with that used in AERO. The parameter settings for each baseline are set according to the specifications detailed in their respective paper or adjusted to the optimum. Given that different anomaly detection methods may adopt varying criteria for selecting anomaly thresholds, we implement the POT method across all the methods to ensure a fair comparison. In POT, we set the initial threshold $level=0.99$ and desired probability $q=0.001$ uniformly for all methods on all datasets. 
% unless otherwise stated
% SR,https://paperswithcode.com/paper/time-series-anomaly-detection-service-at\#code. SPOT,https://github.com/Amossys-team/SPOT. FluxEV,https://github.com/jlidw/FluxEV. Donut,https://github.com/korepwx/donut. OmniAnomaly,https://github.com/smallcowbaby/OmniAnomaly. Anomaly-Transformer,https://github.com/thuml/Anomaly-Transformer. TranAD https://github.com/imperial-qore/TranAD, GDN https://github.com/d-ailin/GDN, ESG https://github.com/LiuZH-19/ESG, TimesNet https://github.com/thuml/TimesNet

% We implement our method and its variants in PyTorch 1.11 with CUDA 12.2. We train them on a server with Intel(R) Xeon(R) Gold 5122 CPU @ 3.60GHz and 1 NVIDIA TITAN Xp graphics card. 
For the model training, we use the Adam optimizer with an initial learning rate of 0.001. We set the length of the sliding window $W$ as 200 and the short window size $\omega$ as 60. We set the number of layers and the number of heads in the Transformer as 1 and 4, respectively. The maximum number of epochs for training is set to be 100 and determined by the early stop mechanism with $patience=5$.
% \textcolor{red}{The number of Transformer layers is 1. We set the dimension of hidden states $d_m$ as 8 and the number of heads $h$ as 4. The maximum number of epochs for training is set to be 100 and determined by the early stop mechanism with $patience=5$. }

\subsection{Evaluation Metrics}
We use precision (Prec), recall, and F1-Score (F1) over the test datasets to evaluate the performance of all the compared methods: $Prec = \frac{TP}{TP+FP}, Recall = \frac{TP}{TP+FN}, F1 = \frac{2\times Prec\times Recall}{Prec + Recall}$,  where $TP$, $TN$, $FP$, and $FN$ are the numbers of true positives, true negatives, false positives, and false negatives respectively.
% \begin{small}
% \begin{equation}\nonumber
% \begin{aligned}
% Prec &= \frac{TP}{TP+FP} \\
% Recall &= \frac{TP}{TP+FN} \\
% F1 &= \frac{2\times Prec\times Recall}{Prec + Recall}
% \end{aligned}
% \end{equation}
% \end{small}%

% In practice, operators generally do not care about point-wise metrics. If any subset of the anomaly segment is detected, the full segment is triggered successfully. 
% Thus, a point-adjust approach was proposed by \cite{Donut} to calculate the performance, followed by many works \cite{Donut,OmniAnomaly,TranAD}. 
% We adopt the point-adjust way too. If any points in the anomaly segment are detected successfully, we think this segment is detected correctly. 
In line with the previous studies~\cite{Donut, OmniAnomaly, TranAD},  we adopt a point-adjust strategy to calculate the performance of these metrics. This strategy is particularly applied in cases where alarms are preferred to be segment-level.
% This strategy is particularly applied in cases where detected anomalies are not continuous. 

\subsection{Overall Performance (RQ1)}
% Table~\ref{table:simulation_performance} and Table~\ref{table:Astroset_performance} present the performance of all the compared methods in terms of precision, recall, and F1-score on the synthetic datasets and the real-world datasets respectively.
% The results show that AERO gets the best F1-score on all datasets. 

\subsubsection{Results for Synthetic Datasets}
% \textbf{1)Results for Synthetic Datasets}

Table~\ref{table:simulation_performance} presents the performance of precision, recall, and F1-score for all the compared methods on the three synthetic datasets. Based on the results, we can make several observations.

First, it is observed that most methods designed for univariate time series anomaly detection, except for SR, show the most competitive performance on the SyntheticHigh dataset. However, these methods exhibit less effective results on the SyntheticLow dataset, apart from SPOT and TM. 
This trend is in contrast to the performance of AERO, primarily due to the limited capability of these methods in recognizing concurrent noise. Therefore, their performance is more susceptible to anomaly-to-noise ratio (A/N). Specifically, a higher A/N ratio, indicating less concurrent noise, corresponds to improved performance. Besides, among all the univariate time series methods, SR achieves the best precision and F1 score, while SPOT excels in recall. This is because SR tends to make conservative predictions regarding potential anomalies, in contrast to the more aggressive strategy applied in SPOT.

Second, for methods designed for multivariate time series anomaly detection, their performance is heavily influenced by their capability to model the specific properties of astronomical observations. Specifically, TimesNet, which forms the foundation model for time series analysis, serves as the best-performing baseline model in terms of average F1 score due to its robust architecture suited for a wide range of time series data. GNN-based methodologies, particularly GDN and ESG, closely follow in performance. Their performance results from the capability to model explicit variate correlations. In contrast, OmniAnomaly, AnomalyTransFormer, and TranAD exhibit comparatively weak due to their limited ability to accurately model concurrent noise.

%OmniAnomaly is the only model based on VAE and gets a relatively bad result on average. Secondly, AnomalyTransFormer and TranAD are both models based on TransFormer, while they have distinctly different performances: AnomalyTransFormer gets the worst result on average in terms of recall and F1-score,  TranAD gets the best recall and achieves a relatively moderate F1-score.  Thirdly, GDN and ESG are both models based on graph neural networks and achieve similar performance: GDN is slightly better than ESG on average. Finally, as a state-of-the-art model, TimesNet indeed achieves a remarkable performance ranking only second to our AERO.

Third, AERO exhibits enhanced performance with the decrease of anomaly-to-noise ratio (i.e., more concurrent noise present in the SyntheticLow dataset). 
% On the one hand, SyntheticLow has the lowest A/N meaning the most concurrent noise. 
This superior performance can be attributed to the design of AERO, which is specifically tailored to tackle concurrent noise and effectively reduce the number of false positives. As a result, the precision of AERO is greatly enhanced in scenarios of a larger ratio of concurrent noise, like the cases encountered in astronomical observations. As the level of concurrent noise diminishes, the relative advantage of AERO gradually diminishes. 
% (e.g., lower recall in the SyntheticLow dataset).
% On the other hand, SyntheticHigh has the highest A/N meaning the most true anomalies segments, AERO's recall decreases slightly. 
% As a result, the F1-score also decreases. 
% In a nutshell: the lower the A/N, the greater the advantage for our AERO. 
% precision and F1-score
More importantly, AERO outperforms the baselines in terms of all metrics on all three simulation datasets. Besides, AERO achieves the highest average F1-score, with an improvement of up to 8.76\% over the second-best performing baseline. 

\begin{table}[t]
\setlength\tabcolsep{3.0pt}
\begin{center}
\caption{Results on the synthetic datasets in terms of Precision, Recall, and F1-score (\%).}
\vspace{-3mm}
% \small
\footnotesize
% \scriptsize
\begin{tabular}{@{}l|lll|lll|lll@{}}
\toprule
\multirow{2}{*}{Method} & \multicolumn{3}{c}{SyntheticMiddle} & \multicolumn{3}{c}{SyntheticHigh} & \multicolumn{3}{c}{SyntheticLow} \\ \cmidrule(l){2-10} 
% \Xcline{2-5}
% & Prec$\uparrow$
                & Prec    & Recall      & F1
                & Prec    & Recall      & F1 
                & Prec    & Recall      & F1  \\ \cmidrule(r){1-10}
TM        &6.08  &28.98 &10.06  &11.64 &39.13 &17.94 &10.19 &49.38 &16.90 \\
SR        & 73.92 & 79.71 & 76.71   & 84.23 & 67.39 & 74.88 & 72.18 & 50.62 & 59.51 \\
SPOT      & 26.74 & 100.0 & 42.20 & 30.91 & 100.0 & 47.23 & 28.05 & 100.0 & 43.81  \\
FluxEV    & 57.40 & 55.07 & 56.21   & 81.36 & 84.78 & 83.04 & 61.16 & 49.38 & 54.64 \\
Donut     & 61.40 & 50.72 & 55.56   & 78.72 & 53.62 & 63.79 & 43.03 & 25.93 & 32.36 \\
OA & 20.37 & 34.78 & 25.70& 26.86 & 28.26 & 27.54 & 44.54& 38.27& 41.17 \\
AT & 29.76 & 14.49 & 19.49& 90.55 & 50.00 & 64.43 & 14.79 & 24.69 & 18.50  \\
TranAD    & 31.03 & 100.0 & 47.36   & 54.16 & 100.0 & 70.26 & 35.68 & 100.0 & 52.60 \\
GDN       & 89.58 & 79.71 & 84.36   & 86.03 & 50.00 & 63.24 & 87.93 & 62.96 & 73.38  \\
ESG       & 79.55 & 71.01 & 75.04   & 85.80 & 63.04 & 72.68 & 69.02 & 50.62 & 58.40  \\ 
TimesNet  & 83.33 & 71.01 & 76.68   & 88.58 & 100.0 & 93.94 & 86.54 & 100.0 & 92.78  \\
\textbf{AERO} &\textbf{90.79} & \textbf{100.0} & \textbf{95.17}& \textbf{90.67} & \textbf{100.0} & \textbf{95.10} & \textbf{92.68} & \textbf{100.0}  & \textbf{96.20} \\

\bottomrule
% \multicolumn{10}{l}{\small }\\
\end{tabular}
\vspace{-4mm}
%}
\label{table:simulation_performance}
\end{center}
\end{table}

% Besides, while it may not achieve the top rank on one specific dataset, AERO achieves the highest average F1-score, with an improvement of up to 8.823\% over the second-best performing baseline. 

% because GDN is based on a static graph structure globally and ESG is based on an evolutionary graph structure having a more powerful modeling capability.

\subsubsection{Results for Real-world Datasets}

\begin{table}[t]
\setlength\tabcolsep{3.0pt}
\begin{center}
\caption{Results on the real-world datasets in terms of Precision, Recall, and F1-score (\%).}
\vspace{-4mm}
% \small
\footnotesize
% \scriptsize
\begin{tabular}{@{}l|lll|lll|lll@{}}
\toprule
\multirow{2}{*}{Method} & \multicolumn{3}{c}{AstrosetMiddle} & \multicolumn{3}{c}{AstrosetHigh} & \multicolumn{3}{c}{AstrosetLow}\\ \cmidrule(l){2-10} 
% \Xcline{2-5}
% & Prec$\uparrow$
                & Prec    & Recall      & F1
                & Prec    & Recall      & F1
                & Prec    & Recall      & F1  \\ \cmidrule(r){1-10}
TM        &8.03 &22.22 &11.79 &62.06 &55.56 &58.63 &14.05 &50.00 &21.94\\
SR        &76.21 &100.0 &86.50   & 74.20 & 100.0 & 85.19    & 82.96 &91.67 &87.09\\
SPOT      &38.43 &100.0 &55.52   & 28.11 & 100.0 & 43.89    & 29.18 &\textbf{100.0}&45.18 \\
FluxEV    &35.65 &22.23 &27.38   & 69.00 & 100.0 & 81.66    & 65.79 &78.57 &71.61\\
Donut     &35.27 &22.23 &27.27   & 70.23 & 100.0 & 82.51    & 81.08 &66.18 &72.87 \\
OA        &41.93 &22.23 &29.05   & 64.10 & 55.56 & 59.52    & 86.37 &75.00 &80.28\\
AT        &68.97 &77.78 &73.11 & 55.89 & 44.44 & 49.51 & 55.76 &25.00 &34.52\\
TranAD    &06.47 &22.23 &10.03   & 11.61 & 44.44 & 18.42    & 41.61 &92.86 &57.47\\
GDN       &79.72 &100.0 &88.71   & 64.94 & 55.56 & 59.88    & 69.20 &33.33 &44.99\\
ESG       &40.24 &22.23 &28.63   & 57.47 & 55.56 & 56.50    & 68.18 &42.86 &52.63\\ 
TimesNet  &41.15 &22.23 &28.86   & 68.09 & 55.56 & 61.19    & 85.54 &91.67 &88.50\\
\textbf{AERO} &\textbf{80.72} &\textbf{100.0} &\textbf{89.33} & \textbf{75.36} & \textbf{100.0} & \textbf{85.95} & \textbf{89.00} &91.67 &\textbf{90.31}\\

\bottomrule
% \multicolumn{10}{l}{\small }\\
\end{tabular}
%}
\label{table:Astroset_performance}
\end{center}
\vspace{-5mm}
\end{table}

Table~\ref{table:Astroset_performance} presents the performance of precision, recall, and F1-score for all the compared methods on the three real-world datasets. Based on the results, we can make several observations.

% Since true anomalies and concurrent noise in the real-world datasets become more complex, the results are not as simple and direct viewing as the synthetic datasets. 
% % So we analyze univariate methods and multivariate methods together. 
% Among the ten baselines, three of them perform best on AstrosetHigh, and four of them get the best performance on AstrosetLow. Further to analysis, among the six multivariate methods, 50\% of them(OmniAnomaly, TranAD, and TimesNet) perform best on the AstrosetLow, showing multivariate methods can capture the relationship among concurrent noise to some extent. Among the four univariate methods, 50\% of them(FluxEV and Donut) perform best on the AstrosetHigh, showing univariate methods tend to perform better on datasets with higher A/N where concurrent noise has less impact.

Among all baselines, SR achieves the best overall performance. The adaptability of the spectral residual approach in SR makes it effective for different anomaly types in real-world datasets. 
Template Matching performs the worst due to the limitations of pre-defined and fixed templates.
Consistent with its performance on synthetic datasets, SPOT achieves the highest recall but quite low precision. This observation again demonstrates its tendency to predict a higher number of anomalies, leading to numerous false alarms. FluxEV, an improvement of SPOT, shows a more balanced precision and recall. Building on the VAE model, Dount and OmniAnomaly produce similar results. However, their performance drops significantly on the AstrosetMiddle dataset, as these methods struggle to capture anomalies consisting of long continuous segments. TranAD, facing similar constraints and being more sensitive to minor fluctuations,  exhibits much worse performance compared to AnomalyTransformer in real-world datasets, in contrast to the results on synthetic datasets. Moreover, GDN demonstrates capability in identifying anomalies of long segments on the AstrosetMiddle dataset. Conversely, TimesNet is effective at detecting anomalies of relatively short time spans on the AstrosetLow dataset.

AERO achieves the best result on the AstrosetsLow dataset and the worst result on the AstrosetsHigh dataset for its own performance. This pattern aligns with the trends observed in the synthetic datasets, particularly in relation to the anomaly-to-noise ratio (A/N). Given AERO's effectiveness in modeling concurrent noise, its strengths become more pronounced in scenarios where the A/N ratio is lower.
% Since AERO is effective at modeling concurrent noise, its advantages are more prominent when the A/N is lower showing concurrent noise occupies a relatively more dominant position.
% In a word, AERO performs best on a dataset with the lowest A/N while worst on a dataset with the highest A/N. 
Furthermore, In comparison with the baselines, AERO surpasses them on all three real-world datasets in most metrics, except for the recall metric on the AstrosetsLow dataset. The most notable strength of AERO lies in its superior precision that exceeds all the baselines. It achieves an average improvement of 5.02\%  over the best baseline. This advantage in precision also contributes to an enhancement in F1-score, with an improvement of up to 2.63\%. The experimental results indicate that AERO is effective at dealing with complex real-world scenarios in astronomical observations.

% Though the anomaly rate of AstrosetsHigh is not the highest among the three real-world datasets, the A/N is the highest, showing anomaly occupies a relatively more dominant position in this dataset.
% From the point of view of A/N, 

\subsection{Ablation Study (RQ2)}

To examine the contributions of various components within our method, we conduct experiments by selectively removing different components to observe the impact on the model performance. Specifically, we implement seven variants of the original model to validate the effectiveness of these components. These variants are divided into two categories: three that modify the temporal reconstruction module, and four that modify the concurrent noise reconstruction module. Each model variant is tested on 
% Besides, we conduct ablation studies on 
one synthetic dataset and two real-world datasets. We introduce the details of these model variants below:

\textbf{1) Effect of Temporal Reconstruction Module}
% [leftmargin=*]
\begin{enumerate}[leftmargin=1em,label=\roman*]
    \item \textbf{w/o temporal:} it removes the temporal reconstruction module, retaining only the concurrent noise reconstruction module in the framework.
    % OnlyGraph 
    \item \textbf{w/o univariate input:} the input to the temporal reconstruction module is changed from univariate time series directly to multivariate time series. This adjustment is intended to demonstrate the effectiveness of modeling each variate independently in this module. 
    % TrasMultiGraph
    \item \textbf{w/o short window:} it removes the input from the short window of length $\omega$ in the temporal reconstruction module.
    % and keep the long input window only. 
    % ShortGraph
\end{enumerate}

\textbf{2) Effect of Concurrent Noise Reconstruction Module}

\begin{enumerate}[leftmargin=1em,label=\roman*.]
% TransBasic，TransMulti，TransStatic，TransDynamic
    \item \textbf{w/o concurrent noise}: it removes the concurrent noise reconstruction module while maintaining only the temporal reconstruction module.
    \item \textbf{w/o concurrent noise \& univariate input}: it removes the concurrent noise reconstruction module and changes the input to the temporal reconstruction module as a multivariate time series.
    \item \textbf{w/o window-wise graph (static)}: it applies a static complete graph to model variate correlations rather than the window-wise graph structure learning technique. 
    \item \textbf{w/o window-wise graph (dynamic)}: it uses a dynamic graph structure rather than a window-wise graph structure learning technique. The dynamic graph is learned based on ESG~\cite{ESG} to contain the output of its evolving graph layer. 
    % without further output of the temporal convolution layer, thus avoiding repetitively modeling temporal patterns.
\end{enumerate}

\begin{table}[t]
\setlength\tabcolsep{2.5pt}
\begin{center}
\caption{Results for Ablation Study(as \%).}
\vspace{-2mm}
% \small
\footnotesize
% \scriptsize
% \resizebox*{0.7\linewidth}{!}{
\begin{tabular}{@{}l|lll|lll|lll@{}}
\toprule
                        & \multicolumn{3}{c}{SyntheticMiddle} 
                        & \multicolumn{3}{c}{AstrosetMiddle} 
                        & \multicolumn{3}{c}{AstrosetLow} \\ \cmidrule(r){2-10}
                        % & \multicolumn{3}{c}{Average}
                        
& Prec  & Recall & F1  & Prec & Recall & F1  & Prec & Recall & F1 \\  \cmidrule(r){1-10} 
AERO      & \textbf{90.79}& \textbf{100.0}&\textbf{95.17}  & \textbf{80.72} & \textbf{100.0} & \textbf{89.33}  & \textbf{89.00} &\textbf{91.67}  & \textbf{90.31}  \\ 
\cmidrule(r){1-10}
% \multirow{3}{*}{Module1} 
1) i &43.75  &20.29 &27.72   & 70.21 &77.78 & 73.80  & 84.43 & 75.00 & 79.44 \\
1) ii &62.50  &28.99 &39.60 & 39.30 &22.22 & 28.39  & 87.04 & 75.00 & 80.57\\
1) iii   &59.52  &36.23 &45.05 & 76.27 &100.0& 86.54  & 83.33 & 57.14 & 67.80 \\
\cmidrule(r){1-10}
% \multirow{3}{*}{Module2} 
2) i  &88.69  & 100.0 & 94.00  & 77.12 &100.0  & 87.08  & 29.59 & 08.33 & 13.00 \\
2) ii &80.80  & 100.0 & 89.38  & 76.34 &100.0  & 86.58  & 87.04 & 75.00 &80.57 \\
2) iii &74.70  &71.01 & 72.81  & 74.07 &100.0  & 85.11  & 86.20 & 91.66 &88.85 \\
2) iv  &83.54  & 100.0 & 91.03  & 73.49 &100.0  & 84.79  & 39.99& 58.33 &47.45\\ 
\bottomrule
\end{tabular}
\vspace{-4mm}
%}
\label{table:Ablation}
\end{center}
\end{table}

The results for different model variants are presented in Table \ref{table:Ablation}. Based on the results, we can observe that the removal of different components from the framework leads to the decline of all the metrics. This demonstrates the contributions of each component to the model performance. Notably, the impact of specific components varies across different datasets. For example, the temporal reconstruction module serves as the most influential factor in the performance of the SyntheticMiddle dataset. For the AstrosetsMiddle and AstrosetsLow datasets, the pivotal roles shift to the univariate input and the concurrent noise reconstruction module. 
% This shifts to the univariate input and the concurrent noise reconstruction module for the AstrosetsMiddle and the AstrosetsLow datasets, respectively.

% On average, the worst variant is w/o univariate input. Substituting multivariate input for univariate input has the highest performance drop of nearly 45.941\% in terms of the F1-score. This result demonstrates the importance of modeling temporal patterns first and independently.

% or remove the short input window in the temporal reconstruction module (w/o short window),
% replace the univariate input with the multivariate one (w/o univariate input)
On average, three model variants that either replace the univariate input with the multivariate one (w/o univariate input), omit the temporal construction module (w/o temporal) or remove the concurrent noise reconstruction module (w/o concurrent noise), produce the most serious effects on descending order. Specifically, these modifications result in a substantial decrease in F1-score by 45.94\%, 34.15\%, and 29.38\% respectively. It is interesting to observe cases where the performance completely collapses when these critical properties are not well tackled due to the above components. This finding highlights the importance of simultaneously addressing both variate independence and the concurrent noise properties, which are unique and pivotal in the context of astronomical observations. 
% considering univariate temporal patterns and the relationship of multivariate concurrent noise together aids anomaly detection performance. Combining Long and short input windows can have better performance.

Moreover, compared to the proposed window-wise graph structure learning technique, the adoption of a static graph or the implementation of dynamic graph structures learning leads to a decrease in F1-score by 10.20\% and 18.76\% respectively. This result demonstrates that while all three methods aim to address concurrent noise, window-wise graph structure learning emerges as the most effective approach. The superior performance is attributed to its more reasonable prior assumption, which effectively models the characteristics of spatial and temporal randomness inherent in concurrent noise. 

% Using a static complete graph is a little better than a dynamic graph because two real-world have concurrent noise at every dimension. While on the SyntheticMiddle, using dynamic graphs is better than a static complete graph. 

% Among all variants, w/o module2 and w/o univariate input achieves the least amount of decline. However, it still leads to a 6.65\% drop in terms of F1-score on average.

\subsection{Model Efficiency and Scalability (RQ3)}
\begin{figure}[ht]
\centerline{\includegraphics[width=0.48\textwidth]{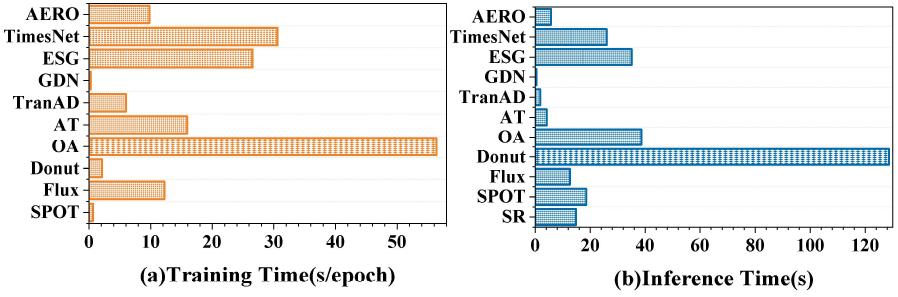}} % Reduce the figure size so that it is slightly narrower than the column.
\vspace{-3mm}
\caption{Results for model efficiency}
\label{fig:traintesttime}
\vspace{-4mm}
\end{figure}

We conduct further experiments to evaluate the model efficiency in terms of training and testing time for all the compared methods. Note that SR is excluded from the analysis since this method does not involve the learning process. The results on the SyntheticMiddle dataset are reported in Fig.~\ref{fig:traintesttime}, and similar trends can be observed on other datasets. For the training stage, it can be observed that OmniAnomaly requires the longest training time per epoch as it utilizes GRU which sequentially processes the data points at each step, whereas GDN is the most time-efficient due to the efficiency of the GNN model utilized in this method. The proposed AERO model, while not the fastest, demonstrated comparable efficiency to these models. Despite its calculation of a distinct graph structure for every sliding window, the number of parameters remains modest to be time-efficient for training. In the testing phase, the trends are similar to those of the training stage: GDN is still the most efficient. Notably, AERO demonstrates competitive efficiency in the testing phase as well.
These results indicate that the efficiency of our proposed AERO is competitive in both the training and testing stages. In this case, AERO can be deployed in online anomaly detection to satisfy the real-time requirement at a relatively low training cost. In addition, AERO strikes a balance between runtime efficiency and anomaly detection performance, making it a practical choice for real-world applications requiring both speed and accuracy.

To study the scalability of AERO, we first analyze its computational complexity. Given the size of long window $W$, the size of short window $\omega$, the number of stars $N$, the dimension of hidden state of Transformer $d_m$, the time complexity is $O(W^2d_m + N\omega ^2)$, which remains the same magnitude as compared to other methods.

\begin{figure}[ht]
\centerline{\includegraphics[width=0.48\textwidth]{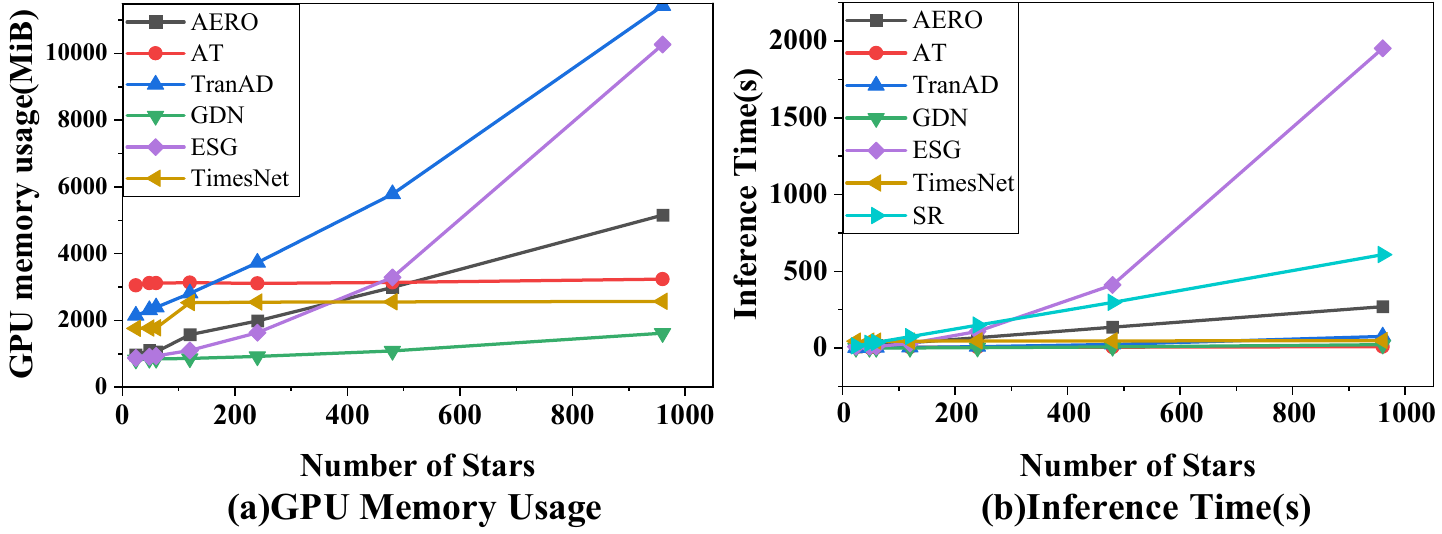}} % Reduce the figure size so that it is slightly narrower than the column.
\vspace{-3mm}
\caption{Results for model scalability}
\label{fig:scalability}
\vspace{-3mm}
\end{figure}

To evaluate its practical applicability, we generated a series of datasets with star numbers ranging from 24 to 960 and tested the GPU memory usage and inference time. Since in practical scenarios, the number of stars in observed images typically does not exceed several hundred (500), we excluded extreme cases such as numbers over 1000. The results for GPU memory usage and inference time are presented in Fig.~\ref{fig:scalability}.
% Since in practical scenarios, the number of stars within a specified celestial region typically does not exceed several hundred (500), we excluded cases that are much beyond typical observational ranges.
For GPU memory usage, we observe a linear increase in AERO, marked by a relatively modest growth rate compared to other baselines, whereas TranAD and ESG demand the highest usage. 
% whereas TranAD, ESG, and TimesNet exhibit the highest inference time
% Furthermore, apart from ESG and SR, which exhibit a significant increase in inference time, the variation in inference time for all other methods is negligible since they treat all variates as one entity. 
For inference time, ESG and SR exhibit a significant increase than AERO, while the increases for other baselines are not significant since they do not compute dynamic correlation matrices.     
% Besides, the inference time increases also linearly for all the methods except ESG. 
Although AERO may not demonstrate the most superior scalability, it achieves comparable memory usage and inference time while exhibiting the highest effectiveness. Thus, AERO meets the requirements of practical applications in the project of scientific discovery.

% We observe a linear increase in GPU memory usage for AERO, marked by a relatively modest growth rate, whereas TranAD, ESG, and TimesNet exhibit the highest inference time. Besides, the inference time increases also linearly for all the methods except ESG. Although AERO may not demonstrate the most superior scalability, it achieves a great trade-off between memory usage and inference time while exhibiting the highest effectiveness. Thus, AERO meets the requirements of practical applications in the project of scientific discovery.

% Since in practical scenarios, the number of stars typically does not exceed 1000, 
% we can conclude that AERO has moderate and acceptable scalability.}

\subsection{Model Analysis (RQ4)}
\begin{figure}[t]
\centerline{\includegraphics[width=0.5\textwidth]{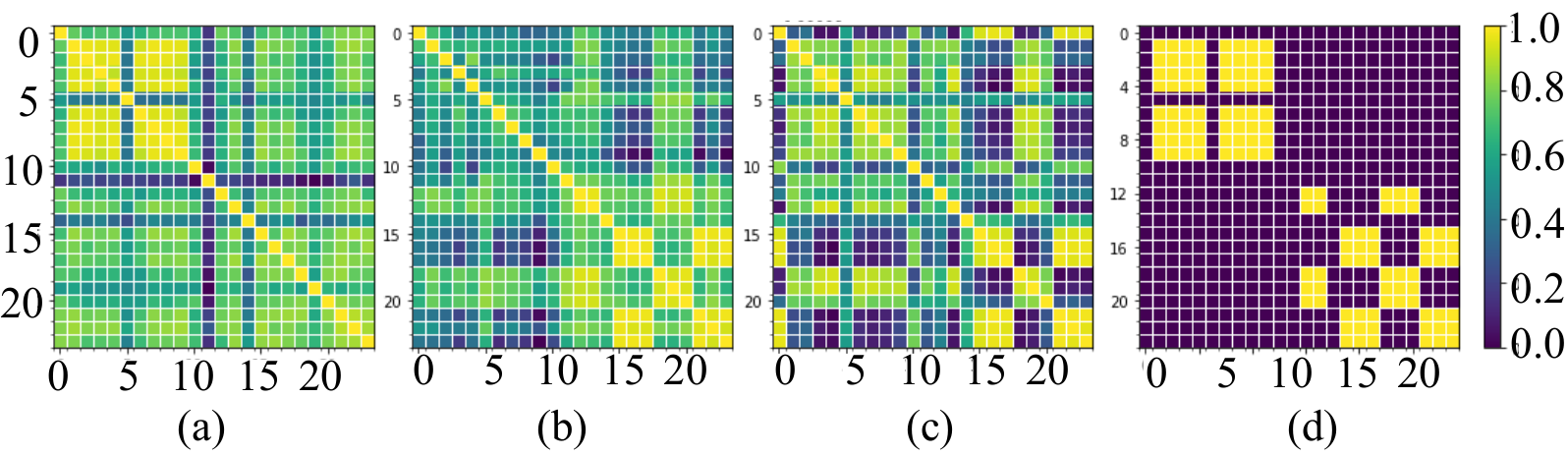}} 
\caption{Visualization of graph structure. (a)-(c) are a series of window-wise graphs from AERO before removing self-loops. They are arranged in temporal order. (d) is the ground truth instances of concurrent noise within the entire time series. 
% (a)-(c) correspond to some parts of (d) respectively meaning they successfully capture the local correlation of concurrent noise.
}
\label{fig:graph}
\vspace{-5mm}
\end{figure}

\textbf{Visualization of the window-wise graph structure.} To further validate the effectiveness of the window-wise graph structure learning technique in capturing concurrent noise, we perform qualitative analysis to visualize the learned graph structure and the ground truth graph constructed based on concurrent noise occurrences in Fig.~\ref{fig:graph}. 
% The rightest graph (d) is a concurrent noise condition known from the ground truth from a whole gable view. 
The yellow dots (i.e., edge weights equal to 1) in Fig.~\ref{fig:graph}(d) represent instances of concurrent noise that affects multiple stars. It is worth noting that these yellow dots include all instances of concurrent noise throughout the entire time series, with each part of noise not necessarily occurring simultaneously. 
Fig.~\ref{fig:graph}(a)-(c) depict samples from the learned window-wise graphs, extracted at different timestamps and arranged in temporal order. We can observe that the module accurately captures instances of concurrent noise within specific time periods. For example, Fig.~\ref{fig:graph}(a) highlights concurrent noise affecting stars 1--4 and 6--9 during early timestamps. Fig.~\ref{fig:graph}(b) captures concurrent noise affecting stars 15--17 and 21--23 as time progresses. Fig.~\ref{fig:graph}(c) also aligns with concurrent noise patterns shown in Fig.~\ref{fig:graph}(d). These results illustrate that the learned windows-wise graph structures effectively capture the actual occurrences of concurrent noise across different timestamps. This ability to accurately capture the true dynamics of concurrent noise in the data validates that window-wise graph structure learning is effective at handling the property in astronomical observations. 
% captures both patterns from the top left and bottom right corner to some extent.

% \begin{figure}[t]
%     \centering
%     \subfigure[concurrent noise] {\label{fig:concurrent_a} \includegraphics[width=0.22\textwidth]{./fig/concurrent_noise.png}}
%     \subfigure[concurrent graph] {\label{fig:concurrent_b} \includegraphics[width=0.22\textwidth]{./fig/concurrent_graph.png}}
%     \caption{Interpretability via graph structure.}
%     \label{fig:graph}
% \end{figure}

\begin{figure}[th]
\centerline{\includegraphics[width=0.46\textwidth]{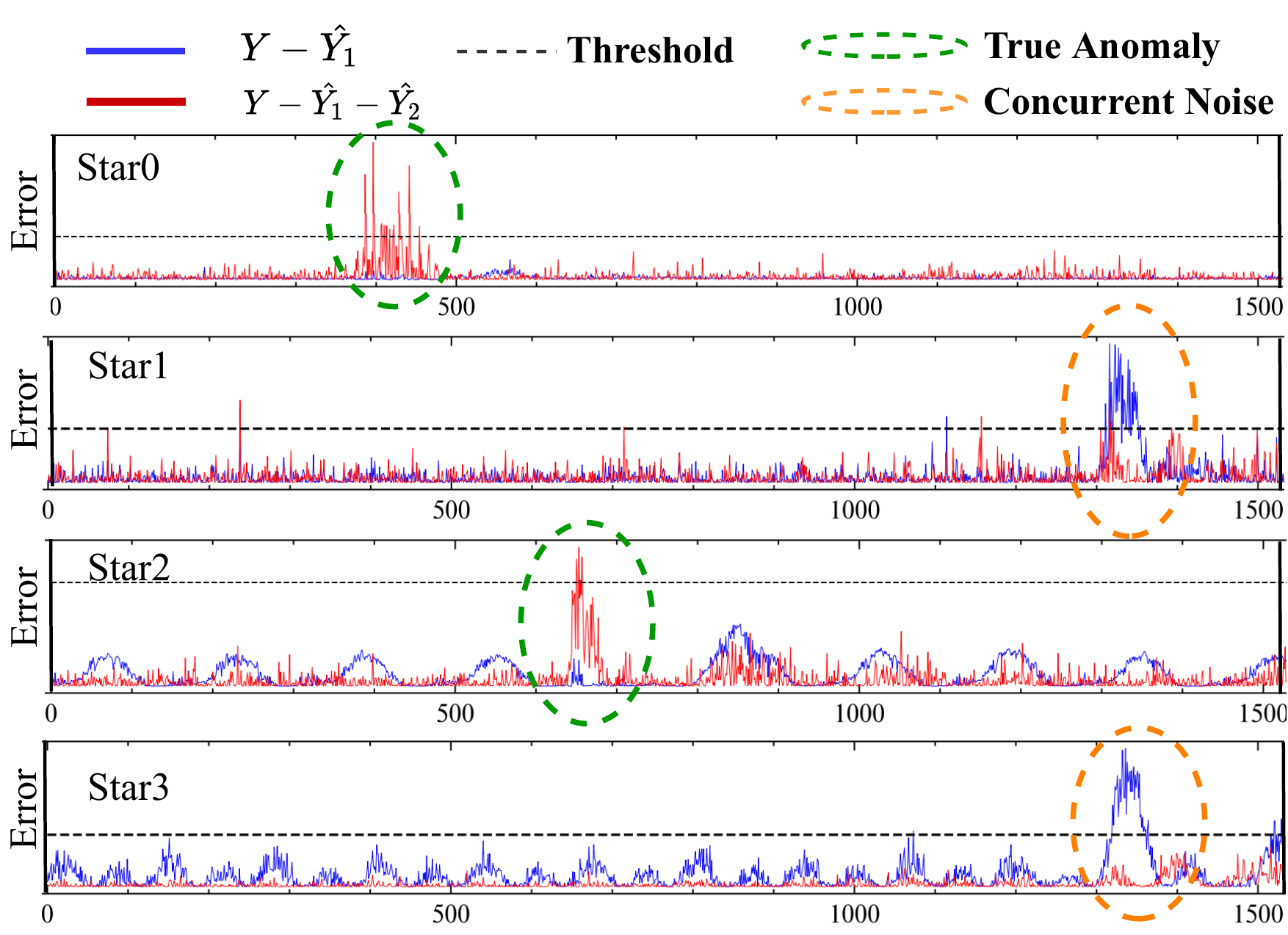}} % Reduce the figure size so that it is slightly narrower than the column.
\caption{Visualization of reconstruction errors. star 0 and star 2 display true anomalies at different timestamps. Concurrent noise occurs on star 1 and star 3 at the same time. Concurrent noise cannot be captured by the temporal reconstruction module but can be filtered out by the concurrent noise reconstruction module.
% This part's final reconstruction errors are lower than those of the temporal reconstruction module. This phenomenon proves AERO can reduce false alarms caused by concurrent noise.
}
\vspace{-3mm}
\label{fig:score}
\vspace{-1.5mm}
\end{figure}

\textbf{Visualization of reconstruction errors.} 
To analyze how each module affects the final anomaly score, we visualize the reconstruction errors $Y-\hat{Y_1}$ from the temporal reconstruction module, together with the final reconstruction error $Y-\hat{Y_1}-\hat{Y_2}$ on several stars in Fig.~\ref{fig:score}.
Among them, star0 and star2 display two true anomalies, while concurrent noise occurs on star1 and star3 at the same time. 
% At the timestamps where true anomaly occur, the final reconstruction error is larger than that of the temporal reconstruction module. While, at the concurrent noise segment, the final reconstruction error is smaller than that of the temporal reconstruction module. 

We can observe that although true anomalies can be successfully detected in the temporal reconstruction module, the segments of concurrent noise are mistakenly classified as anomalies (as indicated by the blue curves surpassing the anomaly threshold), thus leading to false positives. This observation suggests that the temporal reconstruction module, in isolation, is insufficient for addressing concurrent noise without considering the correlations among stars. However, with the incorporation of the concurrent noise reconstruction module, the errors corresponding to these segments are significantly reduced. 
Besides, this module is capable of enlarging the reconstruction errors associated with true anomalies.
%Despite its effectiveness, this module does not affect the reconstruction errors associated with true anomalies. 
Therefore, the combination of these two modules proves to be both reasonable and effective for this task. 

% However, These two opposed phenomena demonstrate that our design for AERO is reasonable and is achieved indeed. 
% Since concurrent noise has correlations with others, these segments can be well reconstructed by the second module of AERO. As a result, the final reconstruction is reduced further. To be more precise, these segments are reconstructed better than using the temporal reconstruction module only. 
% In contrast, at the true anomaly segment, object0 does not have correlations with other objects, so does object2. As a result, the final reconstruction is enlarged by the second module of AERO. Consequently, true anomaly is highlighted.

\subsection{Parameter Sensitivity Analysis (RQ5)}
\begin{figure}[t]
\centerline{\includegraphics[width=0.5\textwidth]{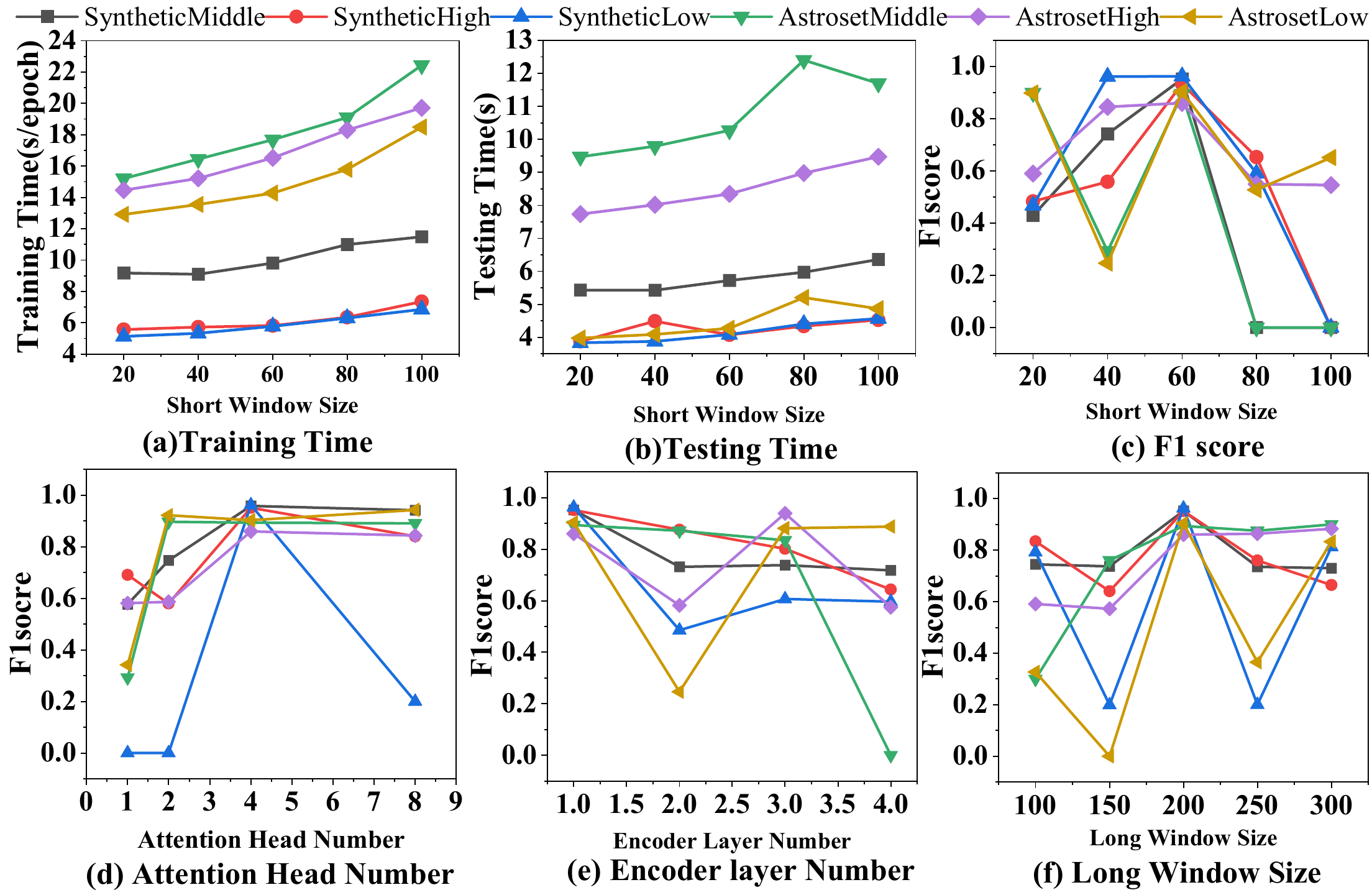}} % Reduce the figure size so that it is slightly narrower than the column.
\caption{Parameter sensitivity analysis in AERO. }
\label{fig:hyper}
\vspace{-7mm}
\end{figure}

% \textbf{Sensitivity of Window Size.}
We study how the value of short window size influences the efficiency and effectiveness of AERO on the six datasets. The results are presented in the upper part of Fig.~\ref{fig:hyper}.
% The results are reported in Fig.~\ref{fig:windowsize}.
% First, for all datasets, the longer the windows, the longer the training time. Second, for most datasets, the longer the windows, the longer the testing time, except for two datasets: AstrosetMiddle and AstrosetLow, which slightly drop when the window size is 100. 
Regarding model efficiency, a general observation is that an increase in short window size corresponds to longer training and testing times (Fig.~\ref{fig:hyper}(a) and (b)). Regarding model effectiveness, while the trends are not uniformly consistent, it is noted that the optimal F1 scores across all datasets are achieved with a short window size of 60 (Fig.~\ref{fig:hyper}(c)). We infer this phenomenon can be attributed to the limitations of both excessively short and long windows: shorter windows may fail to detect smaller anomalies, whereas longer windows might not adequately represent local contextual information. Based on the results, a short window size of 60 achieves a good balance between achieving a high F1 score and maintaining reasonable training and testing time. Consequently, this short window size is selected in our experiments.
% Thirdly, for all datasets, the best F1 scores are obtained when the window size is 60 though the trends are not unified. The optimal F1 scores for most datasets reach the peak at the window size of 60. The F1 scores go up until 60 and then they go down. Excepting for two datasets: AstrosetMiddle and AstrosetLow, they go down first and then go up, and then go down again. But for all datasets, AERO gets the best F1 score when the window size is 60. We infer the reason is that too short windows can not capture small anomalies, While too long windows can not represent the local contextual information well. A window size of 60 gives a reasonable balance between the F1-score and training(testing) times and hence is used in our experiments.
We further study the sensitivity of the other 3 parameters: the head numbers, the number of encoder layers, and the long window size, as depicted in Fig.~\ref{fig:hyper}(d), (e) and (f), respectively. 
% Fig.\ref{fig:hyper}(d) shows model performances with different attention head numbers. 
For the head numbers, the performance is relatively steady under different head numbers. The optimal performance is achieved at 4 in most cases, so we use it in other experiments considering both performance and model complexity.
% For the head numbers, the optimal performance is achieved at 4 in most cases. Besides, the performance is relatively steady under different head numbers. We set the head numbers as 4 in other experiments considering both performance and model complexity.
% Fig.\ref{fig:hyper}(e) shows the performance under different numbers of encoder layers, since many deep neural networks’ performances are affected by the layer number. 
For the number of encoder layers, it can be seen that AERO achieves the best performance with a single encoder layer across all datasets. Therefore, we use only one layer of the encoder, which also makes our model parameter-efficient.
% As shown in Fig.\ref{fig:hyper}(f), the size of the long window is a sensitive parameter. 
For the long window size, the model achieves the best performance at 200 for all datasets, and we set it as the default configuration.

% \begin{figure}[t]
% \centerline{\includegraphics[width=0.5\textwidth]{./fig/windowsize.pdf}} % Reduce the figure size so that it is slightly narrower than the column.
% \caption{Effect of window size. }
% % The training time and testing time grow as the window size gets bigger. But aero achieves the best F1-scores on all datasets when the window size is 60.
% \label{fig:windowsize}
% \vspace{-4mm}
% \end{figure}

\section{Conclusion}

We propose a two-stage anomaly detection model AERO for tackling unique characteristics of variate independence and concurrent noise, in astronomical observations. AERO consists of two modules: the temporal reconstruction module and the concurrent noise reconstruction module. First, AERO uses a Transformer-based module to learn the normal temporal patterns on each variate in consonance with the characteristic of variate independence. Moreover, it devises a novel window-wise graph learning mechanism to equip GNN with the capacity to model random concurrent noise. The extensive experiments on both synthetic datasets and real-world datasets demonstrate the superiority of our method.  
% Compared to the state-of-the-art baseline model, AERO increases F1-scores by up to 8.11\% and 2.56\% on synthetic and real-world datasets respectively.  The smaller the ratio of anomaly to concurrent noise, the more significant the advantage of AERO. This shows that AERO is an ideal choice for anomaly detection in astronomical observations where concurrent noise is a series problem and the true anomaly of value is rare.
In the future, we plan to utilize more scalable and efficient Transformer and GNN variants to model time-series data in more domains.
% In the future, we plan to extend the method of temporal modeling based on more variables of stars and utilize more advanced or larger Transformer variants that are more powerful in modeling sequential data.

% \section*{Acknowledgment}

% This work is supported by the National Science Foundation of China (No.62172423, 91846204, 61941121).

% \section{Acknowledgment}
\section*{Acknowledgement}
This work is supported by the National Natural Science Foundation of China (Grant No: 62172423, U1931133), and the SVOM project, a mission in the Strategic Priority Program on Space Science of CAS.

\balance
\bibliographystyle{IEEEtran}
\bibliography{reference}
\end{document}